\documentclass[10pt,twocolumn,letterpaper]{article}

\usepackage[pagenumbers]{iccv} 

\PassOptionsToPackage{numbers}{natbib}
\usepackage{natbib}
\usepackage{float}
\usepackage{multirow}

\usepackage{siunitx}
\usepackage{amsmath}
\newcommand{\expnumber}[2]{{#1} \times 10^{#2}}
\newcommand{\norm}[1]{\lVert#1\rVert}

\definecolor{LightCyan}{rgb}{0.88,1,1}

\newcommand{\mname}{\emph{CanFields}\xspace}

\definecolor{iccvblue}{rgb}{0.21,0.49,0.74}
\usepackage[pagebackref,breaklinks,colorlinks,allcolors=iccvblue]{hyperref}

\def\paperID{4546} 
\def\confName{ICCV}
\def\confYear{2025}

\title{\mname: Consolidating Diffeomorphic Flows for Non-Rigid 4D Interpolation from Arbitrary-Length Sequences}

\author{Miaowei Wang \quad Changjian Li \quad Amir Vaxman \vspace{2mm}\\
University of Edinburgh
}

\begin{document}
\twocolumn[{%
\renewcommand\twocolumn[1][]{#1}%
\begin{center}
    \centering
    \maketitle
    \captionsetup{type=figure}
    \vspace{-5mm}
    \includegraphics[width=\textwidth, trim=0 0 0 0, clip]{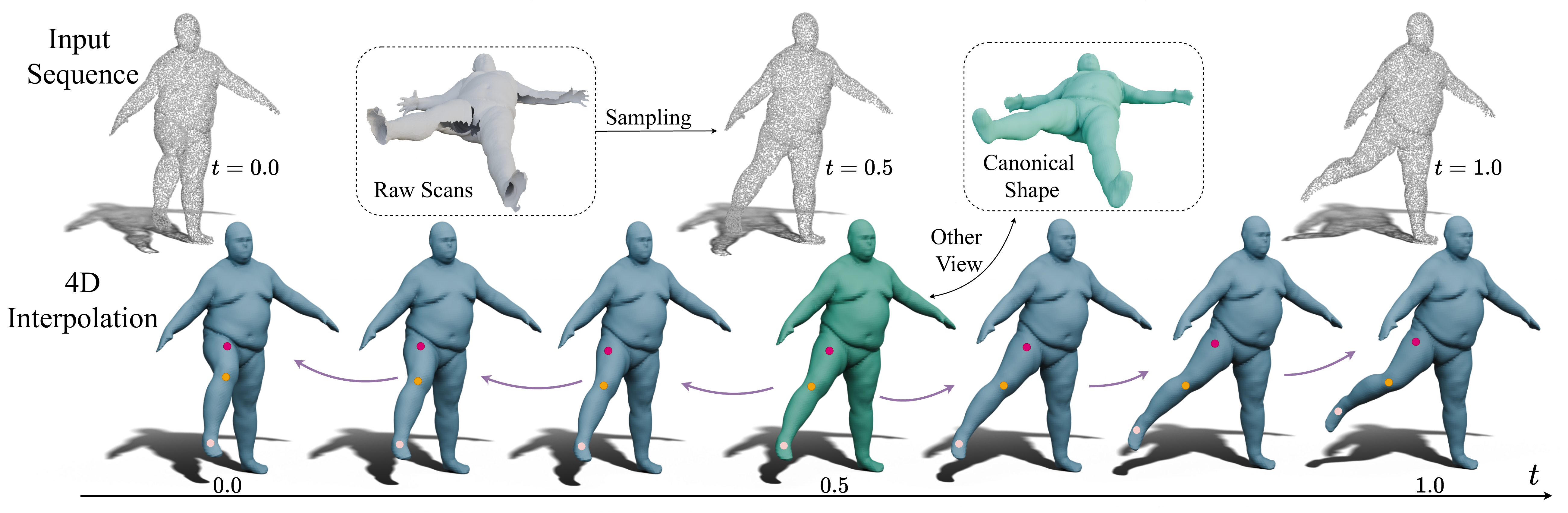}
    \captionof{figure}{\textbf{\textcolor{violet}{\mname}} interpolates the continuous temporal motion of
    a \textcolor[rgb]{0.568,0.776, 0.749}{canonical shape}, by consolidating features from an arbitrary-length  \textcolor{gray}{point cloud sequence} sampled from raw scans. \textcolor[rgb]{0.588,0.451,0.651}{Purple} arrows indicate the optimized flow pushing forward the canonical shape into a corresponding 3D shape at any desired intermediate time. This flow allows for the tracking of any deforming surface point (colored dots) and is robust to scanning defects (in dashed square). See the accompanying video for additional high-quality 4D interpolation results.}
\label{fig:teaser}
\end{center}
}]


\begin{abstract}
We introduce Canonical Consolidation Fields (\mname). This novel method interpolates arbitrary-length sequences of independently sampled 3D point clouds into a unified, continuous, and coherent deforming shape. Unlike prior methods that oversmooth geometry or produce topological and geometric artifacts, \mname optimizes fine-detailed geometry and deformation jointly in an unsupervised fitting with two novel bespoke modules. First, we introduce a dynamic consolidator module that adjusts the input and assigns confidence scores, balancing the optimization of the canonical shape and its motion. Second, we represent the motion as a diffeomorphic flow parameterized by a smooth velocity field. We have validated our robustness and accuracy on more than 50 diverse sequences, demonstrating its superior performance even with missing regions, noisy raw scans, and sparse data. Our project page is at: \url{https://wangmiaowei.github.io/CanFields.github.io/}.
\end{abstract}

\vspace{-5pt}
\begin{flushright}
``Mediation is conflict’s way of looking at itself.''

-- \emph{Jeff Cohen}
\end{flushright}

\section{Introduction}
Accurately reproducing the continuous deformation or motion of scanned physical objects is a fundamental challenge in computer graphics and geometry processing. Motion capture systems, for instance, need to robustly reconstruct moving geometries despite occlusions and partial scans \cite{li2022lidarcap}. Reliable tracking is equally crucial for understanding the natural motions of living beings \cite{Zuffi:CVPR:2024,dong20174d}. In medical imaging, coherent mesh modeling of dynamic objects is vital for assessing volume changes and deformations \cite{cane20184d}.

Typically, scanning devices produce discrete sequences of 3D samples (or \emph{keyframes}) of a moving object.
These keyframes frequently exhibit missing regions from occlusions, measurement noise from imperfect depth estimation or varying acquisition conditions, and complex non-rigid deformation that makes frame-to-frame matching and tracking challenging. Moreover, such sequences can vary significantly in length and lack prior geometric information or markers that establish cross-frame correspondences.

In this paper, we propose a method that takes as input an arbitrary-length sequence of uncorrelated independently-sampled 3D point clouds, and outputs a smooth, \textbf{time-continuous} deformation, faithfully reconstructing the object's geometry and motion. Our approach allows reliable tracking of individual surface points. Importantly, our method imposes no restrictions on the sequence length and does not rely on prior correspondences between keyframes.

\begin{figure}[!t]
\centering
\includegraphics[width=\linewidth]{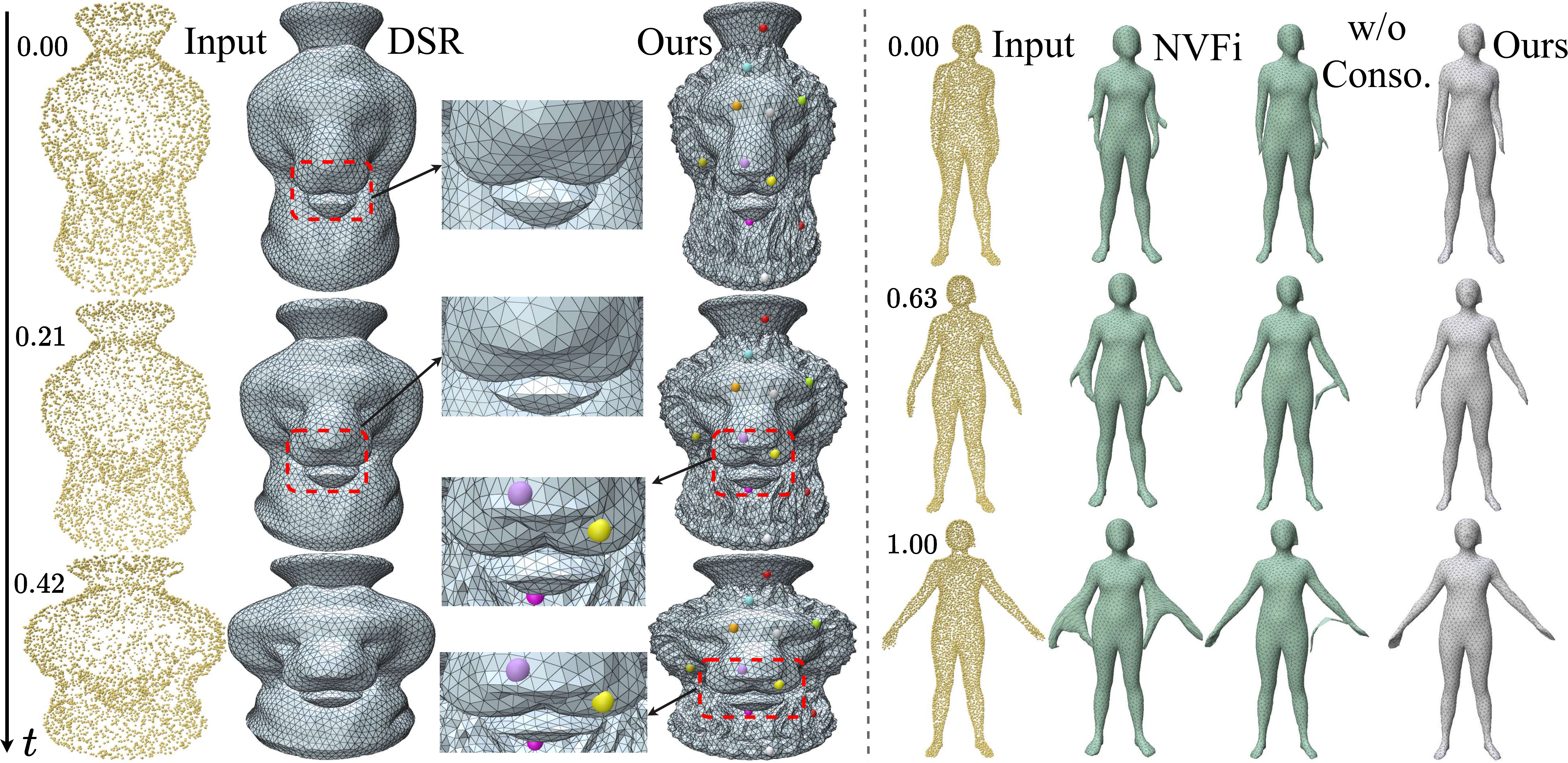}
\vspace{-1em}
\caption{Our method results in a temporally coherent motion of a fine-detailed object, allowing for accurate point tracking. This surpasses 4D implicit methods (\textbf{Left}:  DSR \cite{sun2024dsr}), which oversmooth results and lack time coherency, and flow-based methods (\textbf{Right}: NVFi \cite{li2024nvfi})) that introduce topological artifacts.}
\vspace{-3.6mm}
\label{fig:lion}
\end{figure}
Existing methods addressing this problem typically fall into two categories (\cref{sec:related-work}): (1) \emph{4D implicit} methods~\cite{maoneural,xian2021space,vu2022rfnet} represent dynamic shapes as continuous space-time implicit functions. These methods naturally handle topological changes but often oversmooth geometric details (\cref{fig:lion}, Left) and lack explicit surface point tracking. (2) \emph{Flow-based} methods~\cite{lei2022cadex,tang2021learning,cai2024dynasurfgs} explicitly separate motion from a canonical reference shape, enabling point tracking. However, inaccuracies in the canonical shape can propagate throughout the sequence, causing incorrect topology (\cref{fig:lion}, Right) and compromised geometric fidelity (\cref{fig:4D_interpolation_qualitative,fig:raw_scans}).

Our approach falls into the latter category: we combine an \emph{implicit} fine-detail representation of a canonical static shape with an \emph{explicit}, temporally coherent motion representation between keyframes. However, simultaneously optimizing the canonical shape and motion flow is challenging. The implicit representation often prematurely converges to incorrect geometry, particularly when capturing intricate details (\cref{sec:preliminary}), and motion estimation may fail to aggregate temporal information consistently. This results in a \emph{chimera}-like canonical shape that incorrectly merges features from different frames (\cref{fig:l2_elas_consolidator}). To address this, we introduce a novel \textbf{dynamic consolidator module} (\cref{sec:ioconsolidator}), which incorporates optimizable point displacements to balance canonical shape reconstruction and motion estimation, thus enabling consistent temporal aggregation and accurate recovery of topology and fine geometric detail.
 
\begin{figure*}[!htb]
\centering
\includegraphics[width=\textwidth]{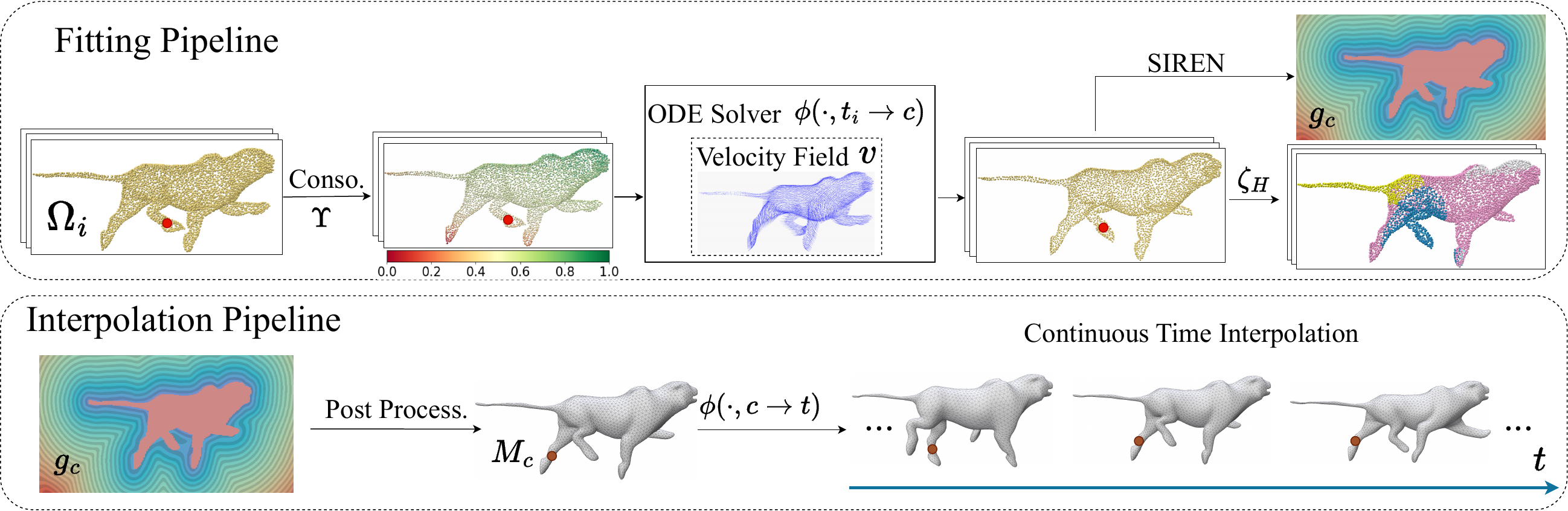}
\vspace{-6mm}
\caption{\textbf{Top (Fitting Phase):} The sampled input point clouds \(\Omega = \{\Omega_i\}\) are processed through our optimizable dynamic consolidator module \(\Upsilon\), reducing variance and estimating confidence. They are then input to the optimizable smooth velocity field \(v\), integrated by an ODE solver to the flow \(\phi\) that transforms points to the canonical time \(c\). An optimizable motion-segmentation network \(\zeta_H\) segments the points for articulation regularization. The reconstructed canonical implicit shape \(g_c\) is encoded with SIREN  \cite{sitzmann2020implicit}. \textbf{Bottom (Interpolation Phase):} We extract a mesh \(M_c\) of \(g_c\) with marching cubes \cite{lorensen1998marching} and flow its vertices to obtain shape \(M_t\) for any target time \(t\). 
}
\vspace{-4mm}
\label{fig:pipeline}
\end{figure*}

Additionally, we propose a \textbf{smooth velocity field} (\cref{sec:low_high_frequence}), integrated into a diffeomorphic motion flow, to capture physically realistic movements of articulated, non-rigid objects without jitter or kinks. This lightweight design also reduces computational complexity in flow integration (\cref{tab:efficiency}). Our overall framework (\cref{fig:pipeline}) consistently outperforms state-of-the-art (SOTA) methods on the task of 4D interpolation from keyframes, particularly in challenging scenarios with missing regions, noisy raw scans, etc. We extensively validate our contributions through ablation studies and experiments on more than 50 diverse sequences. In summary, our key contributions are:
\begin{itemize}
    \item A dynamic consolidator module (\cref{sec:ioconsolidator}) that effectively addresses the joint geometry and motion fitting problem.
    \item A smooth velocity field design (\cref{sec:low_high_frequence}) that ensures physically plausible and coherent motion.
\end{itemize}

\section{Related Work}
\label{sec:related-work}
\paragraph{Shape Evolution} methods~\cite{novello2023neural, eisenberger2021neuromorph, cao2024spectral, liu2022learning, sun2022topology} assume well-reconstructed source and target meshes and learn continuous deformations between them. However, this assumption often fails for real-world inputs, which are frequently fragmented; fitting large discrepancies may yield physically implausible results. Recent progress~\cite{4Deform,sang2025implicit} has been extended to point clouds, yet typically requires prior pairwise correspondence, which we do not need in our work.
\vspace{-12pt}
\paragraph{4D Reconstruction.} Static point-cloud reconstruction is well studied~\cite{carr2001reconstruction, cheng2008survey, kazhdan2006poisson, Hanocka2020point2mesh, mueller2022instant, chen2021multiresolution, shue20233d, ben2022digs, huang2023nksr, williams2021neural, Rui2023GCNO} yet does not extend directly to 4D. Dynamic (4D) reconstruction has explored various input modalities. Video-based methods~\cite{cao2023hexplane, vlasic2008articulated, Mustafa_2016_CVPR, Johnson_2023_CVPR, Lin_2024_CVPR, xian2021space, li2024nvfi, li2021neural, tretschk2024scenerflow, liu2024dynamic} use NeRF~\cite{maoneural, Johnson_2023_CVPR, xian2021space} or Gaussian Splatting~\cite{xu2024grid4d, cai2024dynasurfgs, liu2024dynamic, dgaus} but struggle with geometric consistency, while point clouds~\cite{lei2022cadex, tang2021learning, vu2022rfnet, sharf2008space, anguelov2005scape}, RGBD-SDF~\cite{cong20244drecons, sun2024dsr, atzmon2021augmenting}, and occupancy-based methods~\cite{tang2021learning, vu2022rfnet, cao2024motion2vecsets4dlatentvector, niemeyer2019occupancy, jiang2021learning} better capture geometry yet are vulnerable to noise and missing data. Depending on whether they explicitly model deformation, we classify the methods as either 4D implicit or Flow-based.
\vspace{-15pt}
\paragraph{4D Implicit} methods~\cite{maoneural, xian2021space, vu2022rfnet, cong20244drecons, cao2023hexplane, sun2024dsr, atzmon2021augmenting, li2021neural} represent dynamic sequences as spatiotemporal functions, where each shape is treated as a temporal slice. RFNet-4D~\cite{vu2022rfnet} and \citet{li2021neural} establish correspondences using deformation decoders. SDF Flow~\cite{maoneural} models the scene dynamics via temporal derivatives in PDEs. GenCorres~\cite{yang2023gencorres} and 4DRecons~\cite{cong20244drecons} enforce rigidity constraints via normal directions and extract meshes using marching cubes~\cite{lorensen1998marching}. \citet{atzmon2021augmenting} regularize spatiotemporal derivatives using a Killing loss. Here, we evaluate a state-of-the-art implicit reconstruction method, DSR~\cite{sun2024dsr}, which extends DeepSDF~\cite{park2019deepsdf} to dynamic 4D reconstructions.
\vspace{-15pt}
\paragraph{Flow-based} methods~\cite{lei2022cadex,tang2021learning,cai2024dynasurfgs,sharf2008space,jiang2020shapeflow,groueix20183d,10.1145/3641519.3657520,anguelov2005scape,cao2024motion2vecsets4dlatentvector,Mustafa_2016_CVPR,Johnson_2023_CVPR,Lin_2024_CVPR,li2024nvfi,niemeyer2019occupancy,jiang2021learning,deng2021deformed,InterTransfer24,liu2024dynamic,walker2023explicit} explicitly model the deformation of a canonical shape. This approach is popular in dense correspondence estimation~\cite{bednarik2021temporally,deng2021deformed,InterTransfer24,groueix20183d}, shape-to-shape deformation~\cite{jiang2020shapeflow,groueix20183d,walker2023explicit,zhang2023self}, motion transfer~\cite{muralikrishnan2024temporal,InterTransfer24}, and shape generation~\cite{liu2023exim}. We further categorize these methods into Predefined-Template and Learned-Canonical Flows.
\vspace{-12pt}
\paragraph{Predefined-Template Flow} methods deform a fixed, predefined canonical shape across frames, using standard templates like SMPL for humans~\cite{CAPE:CVPR:20,dfaust:CVPR:2017,Zhang_2017_CVPR,pons2017clothcap}, MANO for hands~\cite{tu2023consistent,lee2023fourierhandflow}, spheres for genus-zero shapes~\cite{walker2023explicit}, or initial-frame reconstructed shapes~\cite{tretschk2024scenerflow}.  This shape may supervise the deformation network directly~\cite{Mustafa_2016_CVPR} or be jointly fitted with it~\cite{dgaus, Lin_2024_CVPR, cao2024motion2vecsets4dlatentvector, 10.1145/3641519.3657520, cai2024dynasurfgs, tang2021learning}. Early approaches like AMA~\cite{vlasic2008articulated} used linear blend skinning to deform articulated human templates, while SCAPE~\cite{anguelov2005scape} extracted rotations from inputs for template deformation.
Recent methods model pose-dependent clothed humans~\cite{CAPE:CVPR:20, pons2017clothcap, Zhang_2017_CVPR}, surpassing earlier minimal clothing setups~\cite{dfaust:CVPR:2017}. 
\vspace{-15pt}
\paragraph{Learned-Canonical Flow} methods, including our work, jointly learn the canonical shape and deformation~\cite{Johnson_2023_CVPR, li2024nvfi, niemeyer2019occupancy, jiang2021learning, saito2021scanimate}. \citet{zhang2023self} introduce hierarchical rigid constraints, while SCANimate~\cite{saito2021scanimate} estimates skinning weights for clothed human shapes. Some approaches leverage nearest-neighbor search in canonical space~\cite{deng2021deformed,InterTransfer24} to establish correspondences. Others, like DG-Mesh~\cite{liu2024dynamic} and \citet{bednarik2021temporally}, combine forward and backward flows enforcing bijective correspondences via cycle consistency.

\noindent \textbf{Neural ODEs}~\cite{chen2018neural} streamline modeling by employing a single velocity network integrated into a diffeomorphic flow, eliminating the need for dual networks. They have demonstrated efficacy across diverse tasks, including 4D reconstruction~\cite{li2024nvfi, niemeyer2019occupancy, jiang2021learning}, physical PDE modeling (ENF~\cite{knigge2024space}), shape deformation (ShapeFlow~\cite{jiang2020shapeflow}), and motion transfer (TemporalJacobian~\cite{muralikrishnan2024temporal}). Numerical integration can hinder convergence during fitting, which our dynamic consolidator mitigates effectively.
\vspace{-12pt}
\paragraph{Dataset Learning} methods~\cite{tang2021learning,cao2024motion2vecsets4dlatentvector,lei2022cadex,merrouche2024combining} extract patterns from large-scale mesh motion datasets. For example, CaDeX~\cite{lei2022cadex} learns deformation embeddings, while Motion2VecSets~\cite{cao2024motion2vecsets4dlatentvector} employs a diffusion model~\cite{zhang20233dshape2vecset} to predict motion latent codes.
However, these methods have a \emph{fixed} number of input-output frames (e.g., 17 frames, dictated by their network architecture), making them inherently incapable of handling arbitrary-length inputs and continuous temporal interpolation. Thus, they are unsuitable as baselines for our task. Furthermore, dataset-based learning methods typically perform well only on motions similar to their training sets and exhibit significant domain gaps on unseen inputs such as raw scans (\cref{fig:raw_scans}).

\section{Temporally-Coherent Reconstruction} \label{sec:preliminary}
Our input is an arbitrary-length temporal sequence of independently sampled 3D point clouds, representing an unknown deforming closed 2-manifold \(S(t)\) with fixed topology and \emph{no given correspondences}. Typically obtained from motion-capture scans, these per-frame point clouds are partial, noisy, and lack temporal coherence. We focus on closed deformable objects (e.g., moving humans or animals), assuming approximately elastic or near-isometric deformation despite imperfections in sampling. 
\vspace{-4.5mm}
\paragraph{Problem Statement.} We consider an input sequence \(\Omega = \{\Omega_i\}\), where each frame \(\Omega_i = (\mathcal{P}_i, \mathcal{N}_i, t_i)\) contains sampled points \(\mathcal{P}_i = \{x_{i,j}\in\mathbb{R}^3\}\), surface normals \(\mathcal{N}_i = \{n_{i,j}\in\mathbb{S}^2\}\), and timestamps \(t_i \in [0,1]\). We compute a continuous implicit function \(f(x,t): \mathbb{R}^3\times\mathbb{R}\rightarrow\mathbb{R}\) whose zero level-set defines the evolving surface \(S_t = \{x\mid f(x,t)=0\}\).

Following~\cite{park2021nerfies,sun2022topology,niemeyer2019occupancy,pumarola2021d}, we decompose the implicit function \( f \) into two components: an implicit \emph{canonical function} \( g_c(x):\mathbb{R}^3 \rightarrow \mathbb{R} \) defined at the canonical time \(c\), with the canonical shape \( S_c = g_c^{-1}(0) \), and a \emph{flow function} \( \phi(x, t \mapsto c) \) mapping a point \( x \) from any time \( t \) back to the canonical time \( c \). The implicit function \( f \) is  defined by the pullback of \( g_c \) through \( \phi \):
\begin{equation}
f(x, t) = g_c \circ \phi(x, t \mapsto c).
\end{equation}
where \( c = 0.5 \) in our experiments. The flow function \(\phi\) is parameterized by a velocity field \( v(x,t):\mathbb{R}^3\times\mathbb{R}\rightarrow\mathbb{R}^3 \), defined as the solution to the initial value problem (IVP): $
\frac{\partial \phi(x,t)}{\partial t}=v(\phi(x,t),t)$
with given initial conditions~\cite{fatunla2014numerical}. Under the Picard–Lindelöf theorem~\cite{siegmund2016generalized}, this ensures that \(\phi\) is diffeomorphic, thus bijective and reversible~\cite{Coddington:1948}. We numerically solve this IVP using a differentiable ODE solver~\cite{torchdiffeq} with the Dormand–Prince (`dopri5') integrator~\cite{dormand1980family}, yielding an approximate diffeomorphism. This naturally allows continuous interpolation without explicitly enforcing cycle consistency~\cite{liu2024dynamic, lei2022cadex}. At evaluation time, we invert the integration direction to obtain the inverse mapping \(\phi(x, c\mapsto t)\), deforming the canonical shape to any target time. The implicit function \(f\), comprising \(g_c\) and \(v\), is jointly optimized using two losses:
\begin{equation} \label{combined_equations}
\begin{aligned}
    \!\!\! E_\text{fit}(i) &= \mathbb{E}_{x_{i,j} \in \mathcal{P}_i} \left( |f(x_{i,j}, t_i)|^2 + \lambda_n \| \nabla f(x_{i,j}, t_i) - n_{i,j} \|^2 \right), \\ 
    E_\text{eik}(i) &= \mathbb{E}_{q \in \mathcal{Q}_i} \left( \| \nabla f(q, t_i) \| - 1 \right)^2.
\end{aligned}
\end{equation}
The first is a fitting term, where \( n_{i,j} \) is the normal at point \( x_{i,j} \), and \( \lambda_n = 0.1 \) balances the point and normal fitting terms. The second is an Eikonal loss to make each $f(x, t_i)$ an SDF. This is fitted on a set of ambient samples \( \mathcal{Q}_i \) per time frame in the unit bounding box, where \( \mathcal{Q}_i \) includes 10K uniformly sampled points and 10K Gaussian samples near the point cloud. The Gaussian is a mixture of \( \mathcal{N}(0, \sigma_1^2 I) \) and \( \mathcal{N}(0, \sigma_2^2 I) \), with \( \sigma_1^2 \) as the distance to the $10^{\text{th}}$ closest point and \( \sigma_2 = 0.3 \) \cite{atzmon2021augmenting}.
\vspace{-3mm}
\paragraph{Canonical Shape Representation.} 
Our goal is to preserve detailed features of the canonical geometry, avoiding oversmoothing common to implicit methods such as DSR~\cite{sun2024dsr} (\cref{fig:lion}, Left). Thus, we represent the canonical shape \( g_c \) as an implicit function encoded by SIREN~\cite{sitzmann2020implicit}. However, directly employing SIREN within existing flow-based methods~\cite{sun2022topology,niemeyer2019occupancy,li2024nvfi} does not yield better detail preservation; paradoxically, it significantly \emph{worsens} results (\cref{fig:othermesthods_use_SIREN}). We clarify this unintuitive outcome next.

\vspace{-3mm}

\begin{figure}[!htb]
\centering
\includegraphics[width=\linewidth]{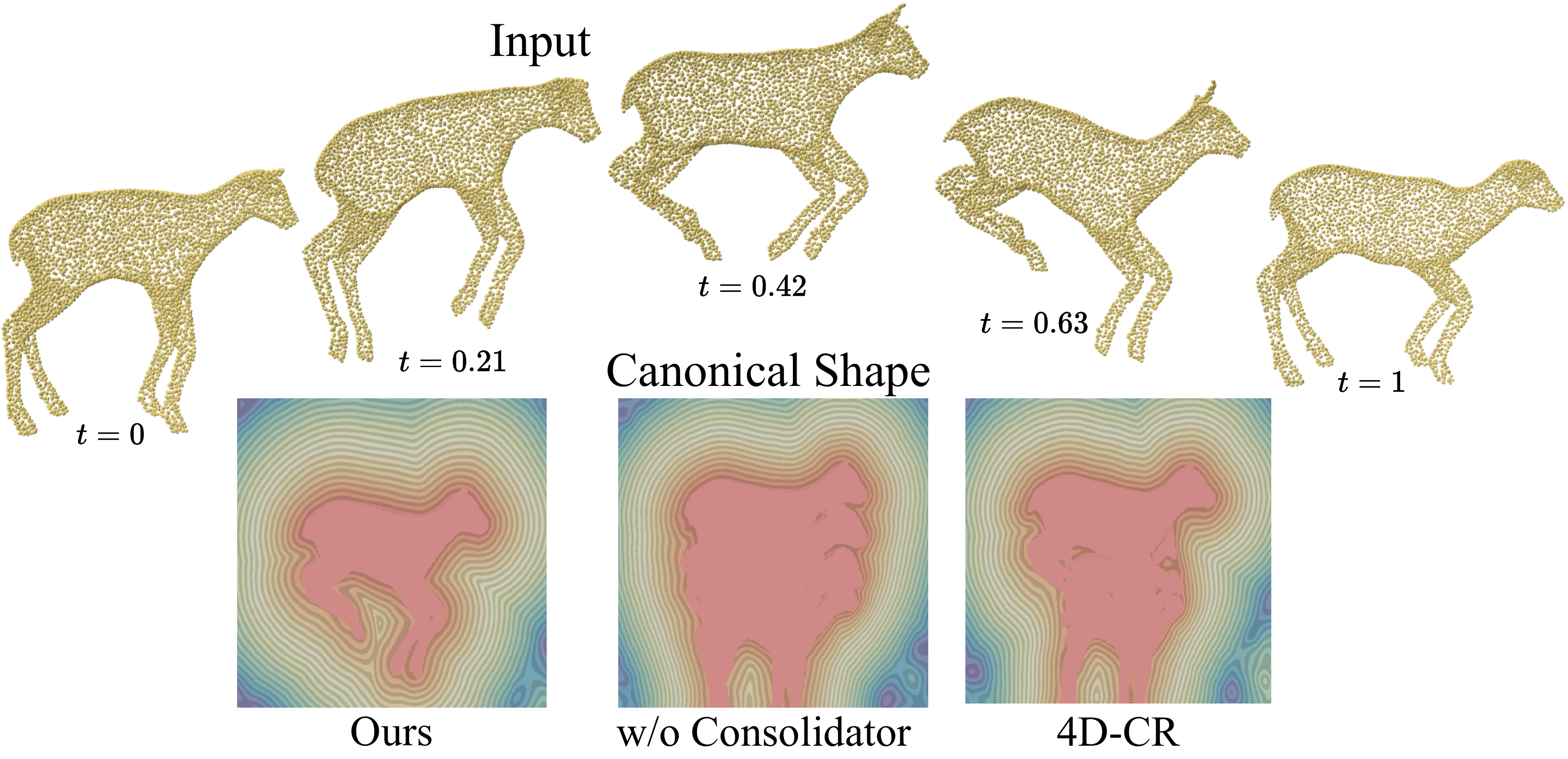}
\vspace{-6mm}
\caption{\textbf{Dynamic consolidator \( \Upsilon \)} effectively resolves conflicts, producing accurate canonical shapes even for rapidly deforming inputs where 4D-CR~\cite{jiang2021learning} fails.}
\vspace{-4mm}
\label{fig:l2_elas_consolidator}
\end{figure}

\paragraph{Geometry \& Flow Fitting Problem.} 
The joint fitting of the velocity field and canonical shape described above is inherently ill-posed~\cite{li2024nvfi,niemeyer2019occupancy,jiang2021learning}, as there are no explicit correspondence or velocity constraints. Small errors in either component propagate throughout the reconstructed 4D geometry, often causing the canonical shape to mistakenly merge features across multiple time frames (\cref{fig:l2_elas_consolidator}). This occurs because numerical integration of the velocity field \( v \) typically fits slower than the canonical implicit function \( g_c \), leading to misaligned or underfitted flows, especially for frames temporally distant from the canonical time. Consequently, the mismatch creates local minima and visual artifacts, particularly in scenes with rapid deformation or noisy inputs. Previous backward-flow methods address this either through manual learning rate tuning~\cite{park2021nerfies,sun2022topology,niemeyer2019occupancy,pumarola2021d} or by emphasizing canonical-time frames as in 4D-CR~\cite{jiang2021learning}, but neither fully resolves issues for rapidly deforming inputs (\cref{fig:l2_elas_consolidator}). Notably, introducing SIREN into these prior methods causes \( g_c \) to fit geometry even faster, exacerbating the mismatch issue and degrading overall performance (\cref{fig:othermesthods_use_SIREN}). In the next section, we propose a novel dynamic consolidator module that mitigates these fitting conflicts by adaptively suppressing erroneous advected features and explicitly evaluating confidence during optimization.

\section{Dynamic Consolidator Module} \label{sec:ioconsolidator}
Inspired by electric transformers that adjust voltage and current to meet an appliance's input requirements, we introduce a learnable transformation of the input operating \emph{before the ODE solver}, coupled with a confidence-based weighting to dynamically reduce the influence of outliers during fitting. Formally, we propose a dynamic consolidator module $\Upsilon: \mathbb{R}^4 \rightarrow \mathbb{R}^4$ with two outputs:
\begin{equation}
\Upsilon: (\eta_{i,j}|\theta_i) \rightarrow (x_{i,j} + \delta_{i,j}, p_{i,j})
\end{equation}
where $\eta_{i,j} = (x_{i,j}, t_i)$ is the space-time coordinate of point $j$ in cloud $\mathcal{P}_i$, and $\theta_i \in \mathbb{R}^{128}$ is a learnable latent code for this cloud. One branch $\delta$ of the consolidator module $\Upsilon$ outputs a spatial deviation function $\delta_{i,j} = \delta(x_{i,j})$ applied to the raw data input. The other branch $p$ provides a confidence score function $p_{i,j} = p(x_{i,j}) \in [0,1]$ for each input point $x_{i,j}$. Consequently, the function $f$ is redefined as:
\begin{equation}
    \widehat{f}(x,t) = g_c \circ \phi(x + \delta, t \mapsto c).
\end{equation}
The deviation $\delta$  homogenizes the input points, to facilitate the consolidation of the points through the flow into the canonical time, for fitting $g_c$. The confidence $p_{i,j}$ measures the reliability at that point, with low confidence expected for outliers. We reformulate the fitting loss $E_\text{fit}(i)$ as follows:
\begin{equation}
\label{raw_fitting_loss}
\begin{split}
    \widehat{E}_{\text{fit}}(i) = &\mathbb{E}_{x_{i,j} \in \mathcal{P}_i}\left[ p_{i,j} \left( |\widehat{f}(x_{i,j},t_i)|^2 + \right. \right. \\
    &\left. \left. \lambda_n \|\nabla \widehat{f}(x_{i,j},t_i) - n_{i,j}\|^2 \right) \right]
\end{split}
\end{equation}
\vspace{-8mm}
\paragraph{Regulating $\Upsilon$.} We regulate $\delta$ both in magnitude and spatial smoothness, with the following two loss terms:
\begin{align}
E_\text{mag}(i) &= \mathbb{E}_{x_{i,j}\in \mathcal{P}_i}||\delta_{i,j}||^2\\
E_\text{var}(i) &= \mathbb{E}_{x_{i,j} \in \mathcal{P}_i}||\nabla \delta_{i,j}||_F^2
\end{align}
We also use a log-likelihood loss to encourage to fit \emph{all} points from the raw consecutive frames:
\begin{equation}
    E_{\text{log}}(i) =  -\mathbb{E}_{x_{i,j} \in \mathcal{P}_i} \log(p_{i,j}),
\end{equation}
which is zero if and only if all confidences are $1$. 
\vspace{-3mm}
\paragraph{Canonical Energy.}
\label{canonic_energy} 
The canonical fitting loss is defined as $E_\text{can}(i) = \lambda_\text{fit}\widehat{E}_\text{fit}(i) + \lambda_\text{eik} E_\text{eik}(i) + \lambda_\text{mag} E_\text{mag}(i) + \lambda_\text{var} E_\text{var}(i) + \lambda_\text{log} E_\text{log}(i)$, where $\lambda$ are hyperparameters.
\vspace{-2mm}
\paragraph{Why dynamic consolidation works.} Our dynamic consolidator module is a variant of \emph{curriculum learning} \cite{bengio2009curriculum,wang2021survey}. To minimize the fitting loss $\widehat{E}_\text{fit}$, the consolidator $\Upsilon$ decreases the confidence scores $p$ and deforms (smooths) the raw point clouds by increasing $\delta$. This in turn produces a smoother motion by focusing primarily on easier inputs with higher $p$. As optimization progresses, the regularization weights $\lambda_\text{mag}$, $\lambda_\text{var}$, and $\lambda_\text{log}$ are gradually increased. This counters the deformation and confidence scores and eventually encourages the model to better fit the original points and motion over time. Thus, it moderates the fitting of $g_c$, allowing $\phi$ to fit more correctly. As the flow is optimized gradually, we get an accurate fitting of the canonical shape, mitigating missing regions and noise. This is apparent in all our results (see \cref{fig:4D_interpolation_qualitative,fig:raw_scans,tab:quan_combined,fig:lion}). We provide the hyperparameter scheduling in Supplementary.
\begin{figure*}[!h]
\centering
\includegraphics[width=\linewidth]{images/qualitative_combined.pdf}
\vspace{-1.5em}
\caption{\textbf{Qualitative Comparisons}. Left: synthetic animal motions from the DeformingThings4D~\cite{li20214dcomplete} dataset. Right: registered real clothed human motions from the CAPE dataset~\cite{pons2017clothcap}. (a) Point Clouds, (b) NVFi~\cite{li2024nvfi}, (c) NDF~\cite{sun2022topology}, (d) OFlow~\cite{niemeyer2019occupancy}, (e) DSR~\cite{sun2024dsr}, (f) Ours. Our approach precisely captures natural temporal deformations with fine details—avoiding over-smoothing and artifacts (see video).}
\vspace{-1.0em}
\label{fig:4D_interpolation_qualitative}
\end{figure*}

\section{Smooth Velocity Field}
\label{sec:low_high_frequence}
\paragraph{Motivation.}  
Our final component is the velocity field \( v \), which integrates to form the flow \(\phi\). We optimize for a motion that is as isometric and elastic as possible. Our velocity representation and optimization eliminate motion jitters and prevent abrupt deformations. Ablation studies in \cref{tab:quan_ablations,fig:low_high_ablation} validate the effectiveness of this strategy.
\subsection{Velocity Architecture}
\begin{figure}[!htb]
    \centering
    \includegraphics[width=\linewidth]{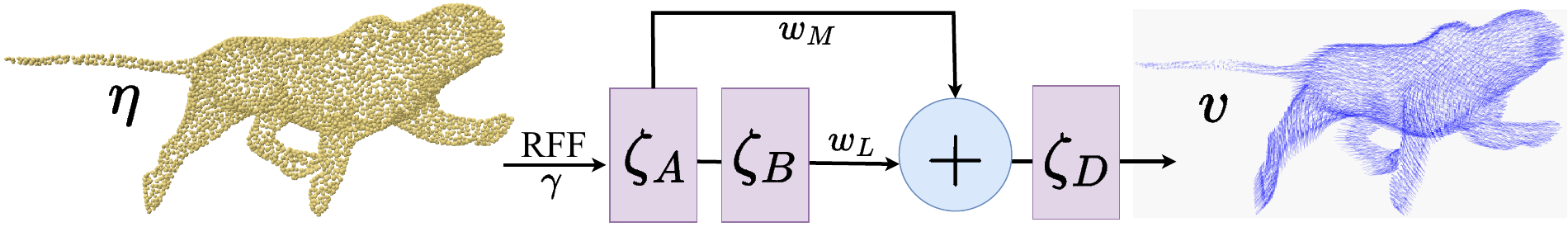}
    \vspace{-7mm}
    \caption{Our smooth velocity field minimal design.}
    \vspace{-4mm}
    \label{fig:velocity-network}
\end{figure}

\begin{figure}[!htb]
    \centering
    \includegraphics[width=\linewidth]{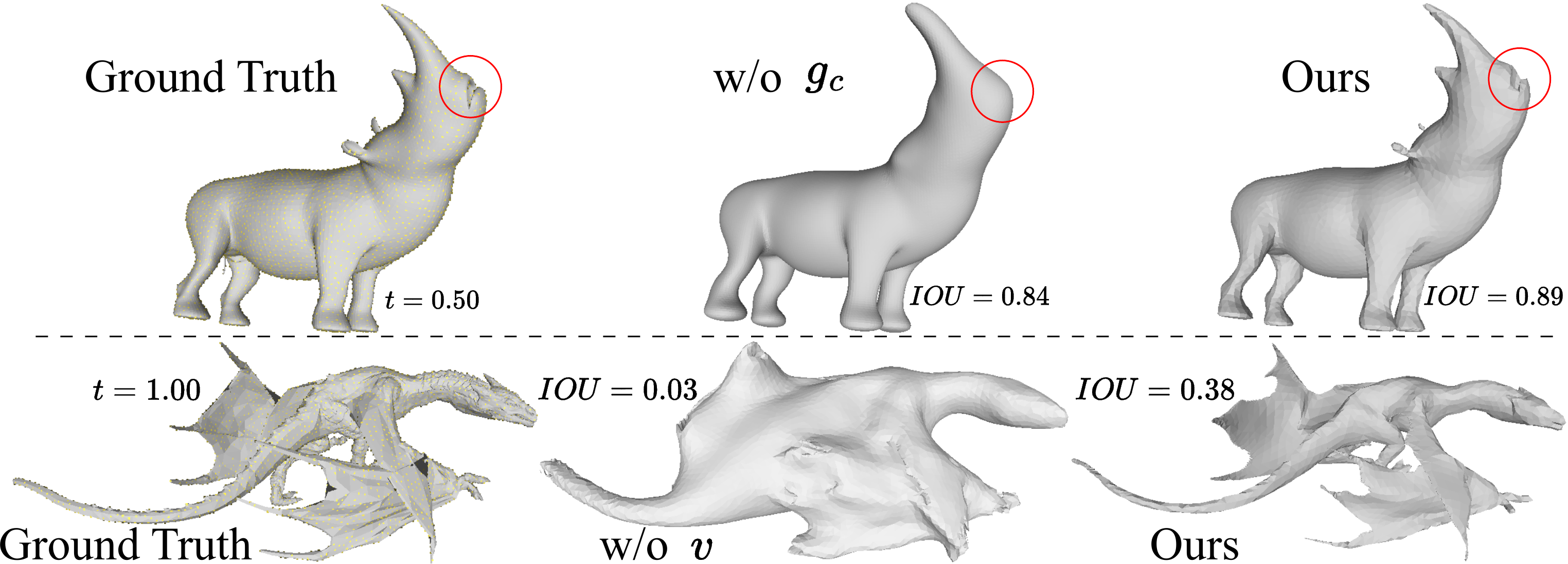}
    \vspace{-0.6cm}
    \caption{\textbf{Geometry and Velocity Ablation.} Our shape representation $g_c$ recovers fine details (mouth in \texttt{rhinoDJ7S\_action3}, Top), while our smooth $v$ prevents kinks and registration errors (\texttt{dragonQKS\_act18}, Bottom). See Sec~\ref{subsec:ablations} for details.}
    \vspace{-0.6cm}
    \label{fig:low_high_ablation}
\end{figure}

Assuming that velocity is smooth in space and time,  we encode Euclidean space-time coordinates into the frequency domain using Random Fourier Features~\cite{walker2023explicit} (\cref{fig:velocity-network}): \(\gamma(\eta) = [\cos(b_1^T\eta), \sin(b_1^T\eta), \dots, \cos(b_{d/2}^T\eta), \sin(b_{d/2}^T\eta)]\), where coefficients \(\{b_i\}\) are sampled from a Gaussian distribution with variance \(\sigma=0.4\). This encoding is processed through two MLP layers, \(\zeta_A\) and \(\zeta_B\), forming a low-pass filter that leverages the inherent smoothness properties of MLPs \cite{liu2022learning,tirer2022kernel} in the composition: \(\zeta_{LM}(\eta) = w_M\zeta_A(\gamma(\eta)) + w_L\zeta_B(\zeta_A(\gamma(\eta)))\), where $w_M=w_L=0.5$; this filters out jitters, kinks and abrupt changes. Ablations for each MLP layer are provided in Supplementary.
Finally, a single-layer coordinate-based decoder maps the filtered Fourier-domain representation into the velocity field \( v \in \mathbb{R}^3 \). This minimal design reduces the parameter search space and significantly \emph{accelerates} ODE integration, resulting in substantial computational savings (\cref{tab:efficiency}).
\subsection{Velocity Regularization}
\label{subsec:velocity-loss-functions}
We assume the velocity field captures physically realistic motions of articulated non-rigid objects, favoring \emph{isometric or elastic} deformations. Unlike direct flow modeling methods~\cite{atzmon2021augmenting}, our explicit velocity formulation requires only first-order spatial ($\nabla$) and temporal ($\frac{\partial}{\partial t}$) derivatives, improving numerical stability and computational efficiency via PyTorch Autograd~\cite{paszke2019pytorch} compared to second-order differentiation~\cite{atzmon2021augmenting}, thus facilitating the following regularizations:

\noindent $\Rightarrow$\textbf{Dirichlet Energy:} $E_\text{diri}(i) = \frac{1}{2} \mathbb{E}_{x_{i,j} \in \mathcal{P}_i} \|{\nabla v(x_{i,j}, t_i)}\|_F^2$ controls spatial variation to force
incompressible flow \cite{sharf2008space}.
\noindent $\Rightarrow$\textbf{Consistent Speed:} $E_\text{speed}(i) = \frac{1}{2} \mathbb{E}_{x_{i,j} \in \mathcal{P}_i} \left\|\frac{\partial v(x_{i,j}, t_i)}{\partial t}\right\|^2$ penalizes sudden movements in unevenly spaced frames and prevents \emph{singularities} (vanishing gradients and zero velocity), especially in sparse or slow-motion frames.
\noindent $\Rightarrow$\textbf{Hierarchical Isometry:} We introduce two complementary regularizations for isometry: the Killing loss (\cref{equa:killing}), enforcing \emph{pointwise} elastic motion, and articulated rigidity (\cref{equa:articulated}) \cite{atzmon2021augmenting, zhang2023self}, encouraging \emph{piecewise} rigid motion.

\textbf{Killing energy} \cite{solomon2011killing, tao2016near, slavcheva2017killingfusion} provides a first-order isometry approximation defined as:
\begin{equation}
E_\text{kill}(i)= \frac{1}{2}\mathbb{E}_{x_{i,j}\in \mathcal{P}_i}\|\nabla v(x_{i,j},t_i) + \nabla v(x_{i,j},t_i)^T\|_F^2.
\label{equa:killing}
\end{equation}

Following previous work \cite{atzmon2021augmenting, zhang2023self}, we adapt a motion-segmentation network \(\zeta_H:\mathbb{R}^3 \rightarrow [0,1]^H\), predicting per-point probabilities \(\zeta_{H,h}(x)\) for point \(x\) belonging to segment \(h\). This yields the \textbf{articulated rigidity} regularization:
\begin{equation}
\begin{split}
E_\text{rigid}(i) = \mathbb{E}_{x_{i,j}\in \mathcal{P}_i}\sum_{h=1}^{H}\zeta_{H,h}(\phi(x_{i,j},t_i\mapsto c))\\
\cdot
\|(R_{h,i}x_{i,j}+\tau_{h,i}) - \phi(x_{i,j},t_i\mapsto c)\|^2,
\end{split}
\label{equa:articulated}
\end{equation}
where rigid transformations \(\{R_{h,i}\in SO(3), \tau_{h,i} \in \mathbb{R}^3\}\) are computed via SVD \cite{belta2002svd} (see Supplementary for details). \textbf{Note:} The only new optimizable module, \(\zeta_H\), operates in the consolidated canonical shape, ensuring \(E_\text{rigid}\) is robust to sparsity, noise, and incomplete inputs over time.

\noindent The total velocity regularization is \(E_\text{velo}(i) = \lambda_\text{rigid} E_\text{rigid}(i) + \lambda_\text{kill} E_\text{kill}(i) + \lambda_\text{diri} E_\text{diri}(i) + \lambda_\text{speed} E_\text{speed}(i)\), where \(\lambda\) are hyperparameters with multi-stage scheduling (see Supplementary) to avoid \emph{local minima}.

\section{Fitting and Interpolation}
\label{sec:training-and-testing}
\paragraph{Fitting Pipeline.} Our fitting architecture (\cref{fig:pipeline}, Top) jointly optimizes the canonical shape \(g_c\) and velocity field \(v\) (\cref{sec:low_high_frequence}). The key components also include the dynamic consolidator module \(\Upsilon\) for coherent fitting (\cref{sec:ioconsolidator}) and a motion-segmentation network \(\zeta_{H}\) for articulated deformation (\cref{equa:articulated}). The \emph{total energy} across the entire input sequence is \(E = \sum_{i} E(i)\) with \(E(i) = E_\text{can}(i) + E_\text{velo}(i)\). 
\vspace{-6mm}
\paragraph{Interpolation Pipeline.} With \(v\) and \(g_c\) fitted, we can produce a continuous 4D interpolation \(M(t)\) for any \(t \in [0,1]\) (\cref{fig:pipeline}, Bottom).  
 Specifically, we discretize the canonical shape \(S_c\) into a mesh \(M_c\) via marching cubes~\cite{lorensen1998marching}, and remesh it~\cite{hoppe1993mesh} to any desired resolution. We then flow vertices of \(M_c\) through \(\phi\) using ODE Solver to produce \(M(t)\).
\section{Results and Evaluation}
Complete implementation details, hyperparameters schedules, additional robustness attacks (e.g., noise, sparsity), and potential applications are provided in Supplementary.
\subsection{4D Interpolation Reconstruction Evaluation}
\begin{figure}[!htb]
\centering
\vspace{-2mm}
\includegraphics[width=\linewidth]{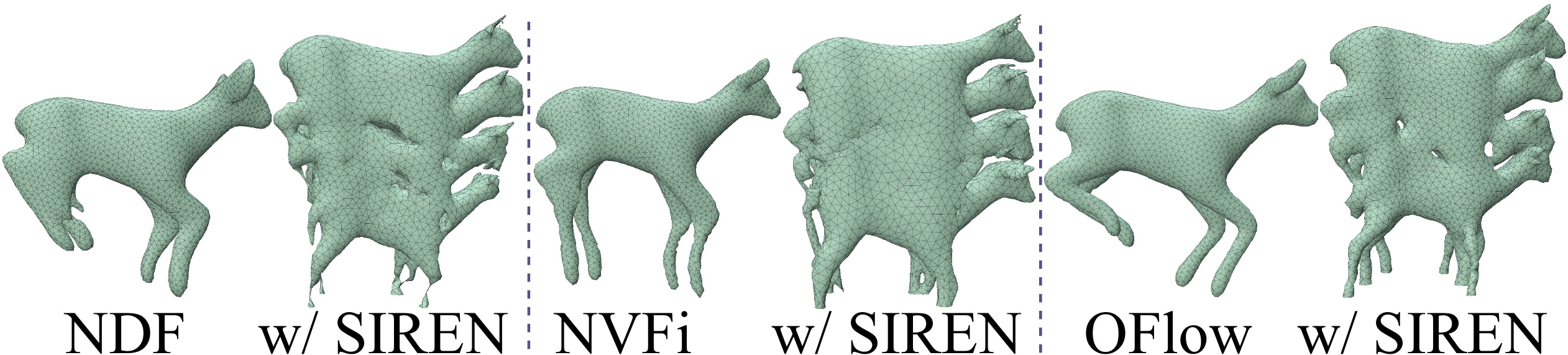}
\vspace{-5mm}
\caption{Using SIREN as we do considerably worsens the canonical shape of other methods, with evidently wrong consolidation.}
\vspace{-7mm}
\label{fig:othermesthods_use_SIREN}
\end{figure}

\paragraph{Benchmark.} We evaluate our method using diverse sequences from two challenging datasets: (1) 26 synthetic motion sequences from the DeformingThings4D dataset~\cite{li20214dcomplete}, capturing a wide range of animal categories and motions; and (2) 8 real human action sequences sampled from the CAPE cloth-human-motion dataset~\cite{pons2017clothcap}. To support evaluation on sequences of arbitrary length, we clip and downsample original sequences to between 10 and 25 frames, in contrast to the fixed 17-frame setup used by dataset-learning methods~\cite{cao2024motion2vecsets4dlatentvector,lei2022cadex}. We spatially sample each frame using Poisson disk sampling~\cite{CCS12}, resulting in sequential point clouds with 4000 points per frame.
\vspace{-4mm}
\paragraph{Beselines.} We compare against SOTA methods that, despite targeting different applications, share key components such as velocity fields. We adapt them for a fair evaluation: (1) \textbf{NDF}~\cite{sun2022topology},  designed for shape evolution with conditional velocity fields and DeepSDF~\cite{park2019deepsdf}; (2) \textbf{DSR}, a 4D implicit method directly predicting SDF values without deformation modeling; (3) \textbf{OFlow}~\cite{niemeyer2019occupancy}, employing ODE-based velocity fields for 4D reconstruction; we adapt it by replacing its shape and identity networks with two optimizable latent codes for per-sequence fitting instead of dataset-level learning; and (4) \textbf{NVFi}~\cite{li2024nvfi}, designed for modeling dynamic 3D scenes from multi-view videos, whose velocity field we incorporate into our adapted OFlow pipeline.
\vspace{-4mm}
\paragraph{Setup.}
Baseline methods represent canonical shapes with DeepSDF~\cite{park2019deepsdf}, whereas our approach uses the more expressive SIREN~\cite{sitzmann2020implicit}. As explained in \cref{sec:preliminary}, directly employing SIREN in baselines leads to premature convergence before velocity fields fully optimize (Fig.~\ref{fig:othermesthods_use_SIREN}). Our consolidator module resolves this, enabling stable SIREN fitting (Fig.~\ref{fig:l2_elas_consolidator}). Thus, we retain DeepSDF for baselines and exclusively adopt SIREN in our method. We set canonical time consistently to $c=0.5$ for all flow-based methods, performing 15,000 fitting iterations, except for OFlow, limited to 10,000 due to its higher computational cost (Table~\ref{tab:efficiency}).
\paragraph{Metrics.} We adopt standard metrics for quantitative evaluations: mean Intersection-over-Union (\textbf{IoU}) \cite{rezatofighi2019generalized}, Chamfer-L1 Distance (\textbf{CD}) \cite{fan2017point}, and Normal Consistency (\textbf{NC}) \cite{zhang20233d}. For thorough assessment, we report both the average and worst-case scores along the temporal dimension. Specifically, we \emph{interpolate 50 frames uniformly} between start ($t=0$) and end time ($t=1$), evaluating each fitted interpolation frame against the ground-truth mesh that is nearest in time, under the common assumption that temporal differences at this granularity are negligible (as in~\cite{sun2024dsr,sun2022topology,zeng2022idea}).
\begin{table}[!htb]
  \centering
  \vspace{-4mm}
    \caption{\textbf{Quantitative Results.} Performance on DeformingThings4D~\cite{li20214dcomplete} and CAPE~\cite{CAPE:CVPR:20} datasets, with best results highlighted.}
      \vspace{-2mm}
    \label{tab:quan_combined}
  \resizebox{\linewidth}{!}{
  \begin{tabular}{c c c c c c c c c c c c c}\hline
    \toprule
     \multicolumn{1}{c}{\multirow{2}{*}{Dataset}} & \multirow{2}{*}{Method} & \multicolumn{2}{c}{IoU (\%)} & \multicolumn{2}{c}{CD ($\times 10^{-4}$)} & \multicolumn{2}{c}{NC ($\times 10^{-2}$)} \\
    \cmidrule(lr){3-4}
    \cmidrule(lr){5-6}
    \cmidrule(lr){7-8}
    \multicolumn{1}{c}{} & & Mean $\uparrow$ & Min $\uparrow$ & Mean $\downarrow$ & Max $\downarrow$ & Mean $\uparrow$ & Min $\uparrow$ \\
    \midrule
    \multirow{6}{*}{\citet{li20214dcomplete}} & NDF~\cite{sun2022topology} & 74.43 & 58.45 & 14.20 & 70.40 & 75.82 & 68.27 \\
    & DSR~\cite{sun2024dsr} & 75.26 & 69.50 & 6.762 & 15.81 & 77.59 & 72.68 \\
    & OFlow~\cite{niemeyer2019occupancy} & 78.47 & 71.56 & 8.012 & 32.28 & 77.17 & 72.37 \\
    & NVFi~\cite{li2024nvfi} & 75.59 & 68.95 & 12.00 & 44.12 & 75.50 & 71.50 \\
    & \cellcolor{LightCyan}\textbf{Ours} & \cellcolor{LightCyan}\textbf{82.84} & \cellcolor{LightCyan}\textbf{74.12} & \cellcolor{LightCyan}\textbf{4.606} & \cellcolor{LightCyan}\textbf{15.28} & \cellcolor{LightCyan}\textbf{78.21} & \cellcolor{LightCyan}\textbf{73.37} \\
    \midrule
    \multirow{5}{*}{\citet{CAPE:CVPR:20}} & NDF~\cite{sun2022topology} & 79.55 & 59.30 & 14.03 & 84.71 & 82.96 & 73.76 \\
    & DSR~\cite{sun2024dsr} & 80.23 & 77.42 & 10.13 & 14.38 & 83.35 & 81.93 \\
    & OFlow~\cite{niemeyer2019occupancy} & 77.15 & 73.38 & 28.55 & 35.00 & 81.61 & 80.08 \\
    & NVFi~\cite{li2024nvfi} & 81.50 & 75.73 & 21.62 & 25.39 & 82.85 & 80.73 \\
    & \cellcolor{LightCyan}\textbf{Ours} & \cellcolor{LightCyan}\textbf{84.17} & \cellcolor{LightCyan}\textbf{80.94} & \cellcolor{LightCyan}\textbf{6.554} & \cellcolor{LightCyan}\textbf{11.23} & \cellcolor{LightCyan}\textbf{84.98} & \cellcolor{LightCyan}\textbf{83.06} \\
    \bottomrule
  \end{tabular}
  }
\end{table}
\vspace{-5.5mm}
\paragraph{Results.}  We quantitatively demonstrate (\cref{tab:quan_combined}) that our method surpasses baselines across all metrics, for both synthetic animal and real human motion datasets. Qualitative comparisons (\cref{fig:4D_interpolation_qualitative}) further highlight our model's temporal coherence and superior detail reconstruction. In challenging, fast-moving sequences (e.g., woman, \cref{fig:4D_interpolation_qualitative}, Right), NDF and NVFi fail to accurately align corresponding parts across frames, whereas OFlow (with a deeper velocity network) and DSR perform better. However, DSR's single SDF network oversmooths results, significantly reducing geometric details in complex scenarios, as illustrated by the elephant (\cref{fig:4D_interpolation_qualitative}, Left) and lion (\cref{fig:lion}, Left).
\subsection{Robustness Attacks}
\begin{figure}[!htb]
\centering
\vspace{-4mm}
\includegraphics[width=\linewidth]{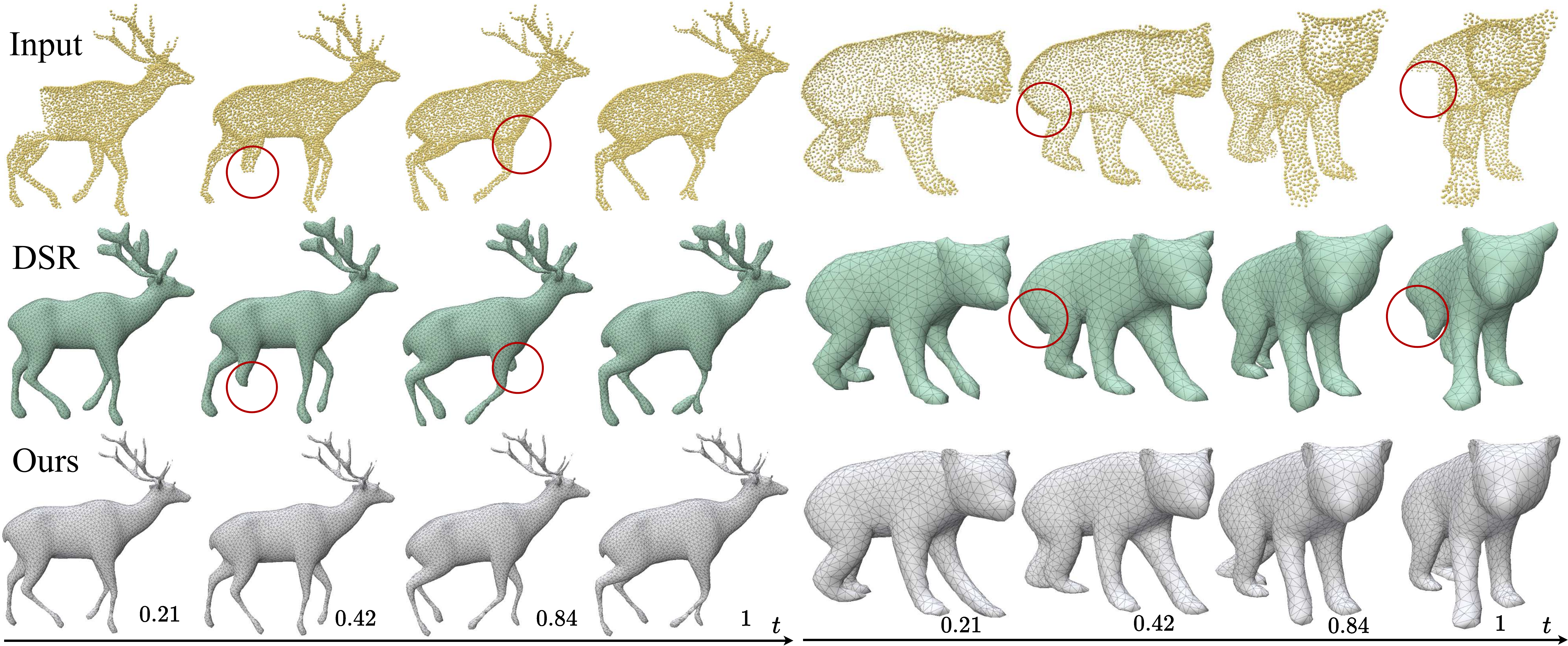}
\caption{\textbf{Missing Regions.} Our method (gray) correctly recovers geometry and deformation, while DSR (green) misses them.}
\label{fig:miss}
\vspace{-8.5mm}
\end{figure}

\begin{figure}[!htb]
\centering
\includegraphics[width=\linewidth]{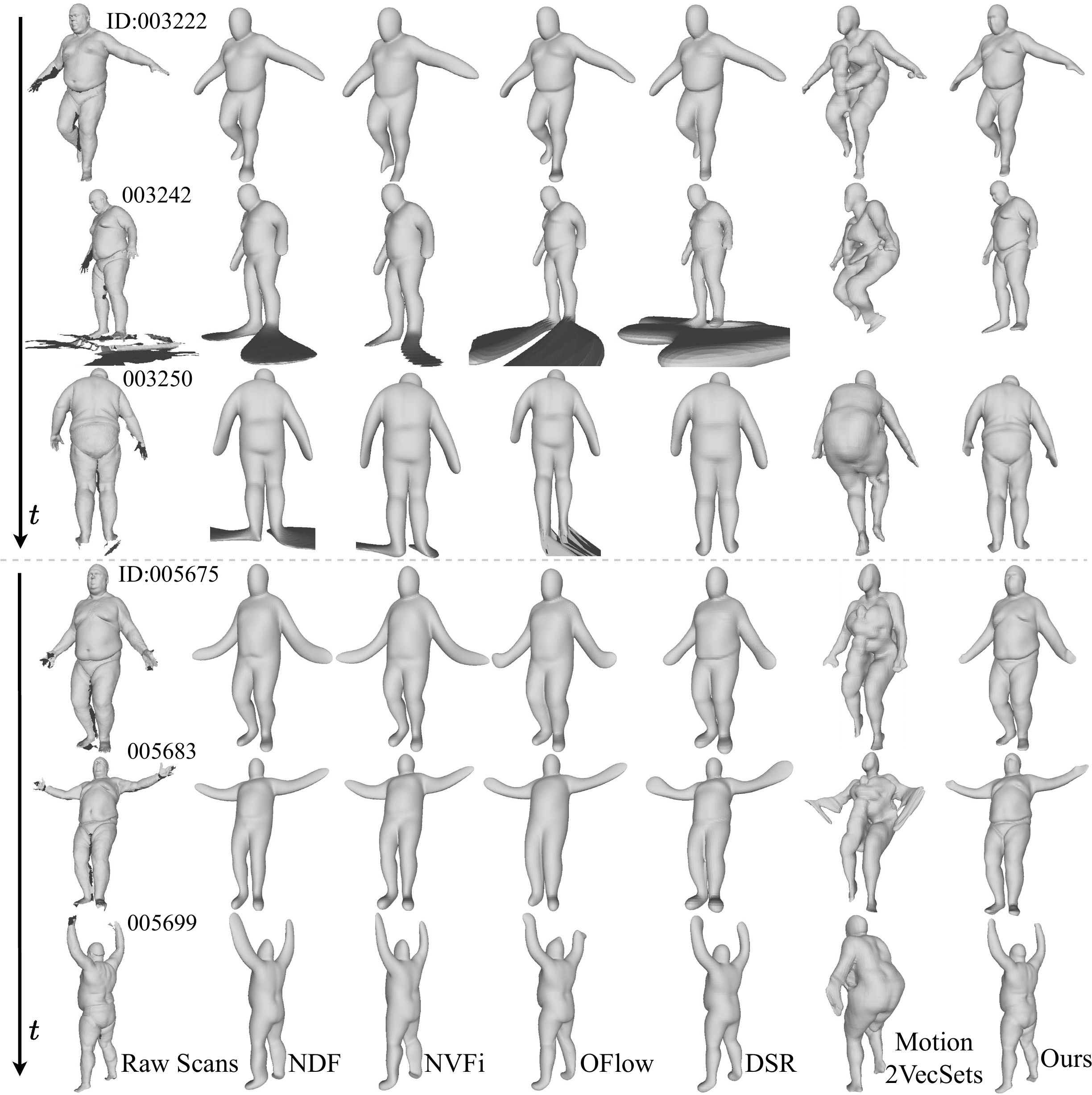}
\vspace{-5mm}
\caption{\textbf{Raw Scans.} Our method significantly reduces artifacts compared to baselines for DFAUST sequences \texttt{one\_leg\_loose} (Top) and \texttt{jumping\_jacks} (Bottom), Subject ID 50002~\cite{dfaust:CVPR:2017}. Cropped visuals shown; see supplementary video for full details.}
\label{fig:raw_scans}
\vspace{-2mm}
\end{figure}

\paragraph{Missing Regions.}
To test our method’s ability to aggregate partial shape information across frames, we manually remove specific regions from each input frame. As illustrated in~\cref{fig:miss}, our approach, leveraging explicit velocity field modeling and regularization, effectively integrates partial details to reconstruct the complete canonical shape. In contrast, DSR fails to recover the missing regions (see quantitative results and video comparisions in Supplementary).
\paragraph{Raw Scans.}The raw scans in DFAUST~\cite{dfaust:CVPR:2017} contain realistic artifacts, including outliers and irrelevant geometry (e.g., ground surfaces),  but lack ground truth meshes; thus, we perform qualitative comparisons.
 We select 17 frames for a fair comparison with pre-trained, dataset-learned SOTA Motion2VecSet~\cite{cao2024motion2vecsets4dlatentvector}. As shown in~\cref{fig:raw_scans}, existing flow-based and implicit methods often struggle with missing regions and outliers, whereas our method consistently produces accurate reconstructions. Motion2VecSet generalizes poorly to unseen raw scans without ground truth for fine-tuning and, due to its diffusion network, reconstructs only a fixed set of discrete frames. In contrast, our method requires no pretraining, performs single-sequence fitting, and naturally supports continuous temporal interpolation.
\subsection{Analysis and Ablation}
\label{subsec:ablations}
\paragraph{Quantitative Ablations.} We quantitatively evaluate our key modules on five diverse sequences from DeformingThings4D dataset in \cref{tab:quan_ablations}: (1) \textbf{Dynamic Consolidator ($\Upsilon$)}. Removing the consolidator module (w/o $\Upsilon$) imbalances the joint fitting of shape and velocity field, producing noticeable artifacts in \cref{fig:l2_elas_consolidator}. (2) \textbf{Canonical Shape Representation ($g_c$)}. Replacing our SIREN-based canonical shape with DeepSDF (w/o fine $g_c$, \cref{fig:low_high_ablation} Top) significantly reduces interpolation accuracy. (3) \textbf{Smooth Velocity Field ($v$)}. Substituting our smooth velocity field with NVFi's formulation (w/o our $v$, \cref{fig:low_high_ablation} Bottom) slightly degrades performance, demonstrating robustness. (4) \textbf{Articulated Rigidity Loss ($E_\text{rigid}$)}. Omitting the articulated rigidity loss (w/o $E_\text{rigid}$) reduces accuracy in modeling articulated motion. Overall, each proposed component contributes meaningfully to accurate and robust 4D interpolation.
\begin{table}[!htb]
  \centering
\caption{\textbf{Quantitative Ablations.} We evaluate the impact of the dynamic consolidator ($\Upsilon$), canonical shape representation (fine $g_c$), smooth velocity field ($v$), and articulated rigidity loss ($E_\text{rigid}$).}
    \label{tab:quan_ablations}
    \vspace{-2mm}
  \resizebox{\linewidth}{!}{
  \begin{tabular}{c c c c c c c c}
    \toprule
   \multirow{2}{*}{Method} & \multicolumn{2}{c}{IoU (\%)} & \multicolumn{2}{c}{CD ($\times 10^{-4}$)} & \multicolumn{2}{c}{NC ($\times 10^{-2}$)} \\
    \cmidrule(lr){2-3} \cmidrule(lr){4-5} \cmidrule(lr){6-7}
    & Mean $\uparrow$ & Min $\uparrow$ & Mean $\downarrow$ & Max $\downarrow$ & Mean $\uparrow$ & Min $\uparrow$ \\
    \midrule
    \cellcolor{LightCyan}\textbf{Default} & \cellcolor{LightCyan}\textbf{83.23} & \cellcolor{LightCyan}\textbf{78.34} & \cellcolor{LightCyan}\textbf{2.193} & \cellcolor{LightCyan}\textbf{7.481} & \cellcolor{LightCyan}\textbf{82.22} & \cellcolor{LightCyan}\textbf{78.49} \\
    w/o $\Upsilon$ & 57.98 & 35.74 & 171.2 & 544.2 & 67.73 & 57.39 \\
    w/o fine $g_c$ & 60.20 & 57.60 & 496.2 & 630.0 & 75.65 & 73.08 \\
    w/o our $v$ & 81.00 & 75.04 & 7.783 & 37.75 & 80.67 & 76.14 \\
    w/o $E_\text{rigid}$ & 73.07 & 65.19 & 662.7 & 2473 & 78.35 & 74.57 \\
    \bottomrule
  \end{tabular}
  }
  \vspace{-3mm}
\end{table}

\begin{figure}[!htb]
\includegraphics[width=1.0\linewidth]{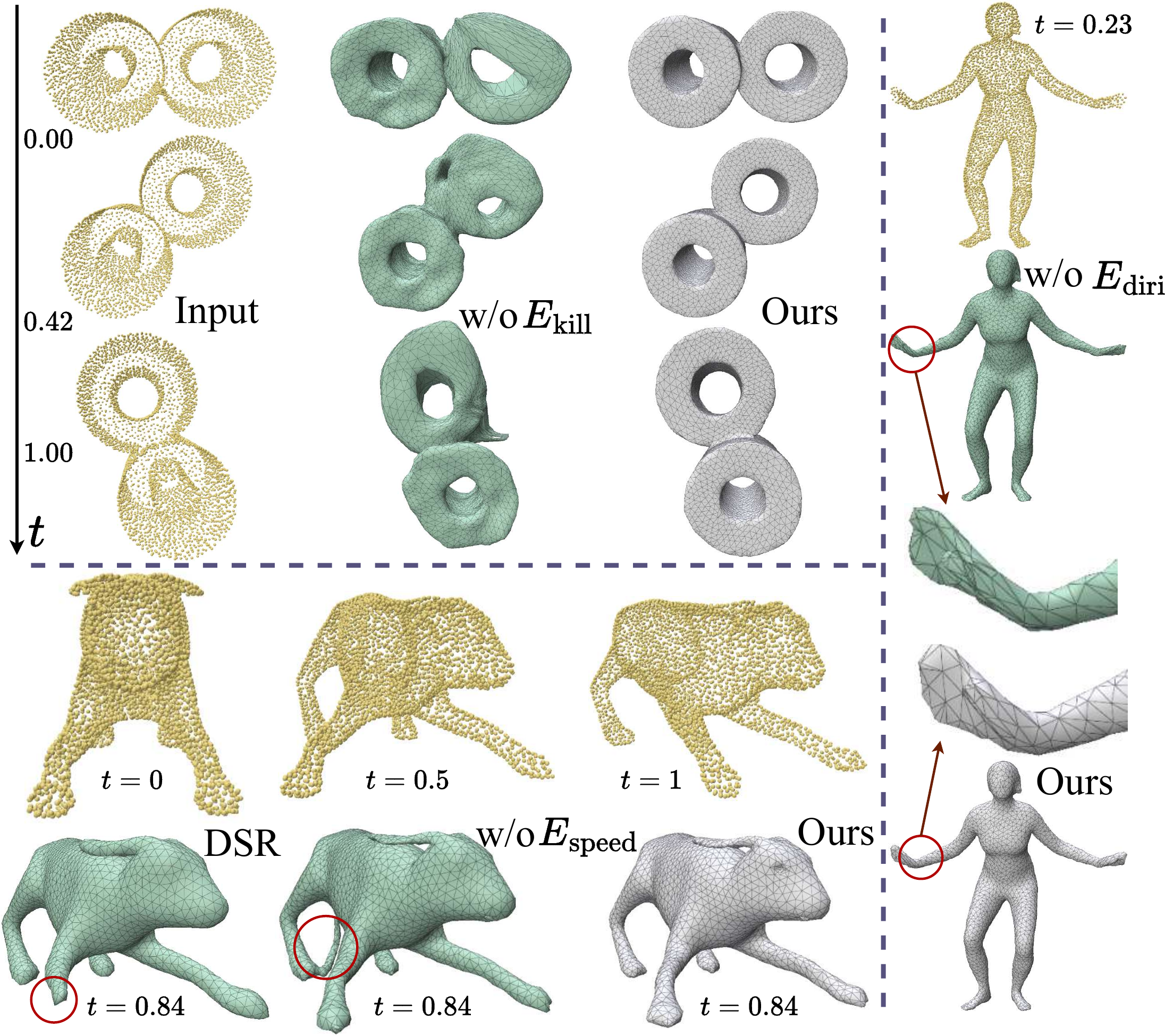}
\vspace{-5mm}
\caption{\textbf{Top}: \(E_{\text{kill}}\) reproduces rigid transformations, minimizing artifacts. \textbf{Right}: \(E_{\text{diri}}\) preserves local shape during deformation (e.g., the right wrist). \textbf{Bottom}: \(E_{\text{speed}}\) facilitates smooth motion even with only three input frames. Artifacts are highlighted in red.}
\vspace{-4mm}
\label{fig:kill_diri}
\end{figure}

\vspace{-4mm}
\paragraph{Qualitative Ablations.} We further evaluate the importance of our regularization terms in~\cref{fig:kill_diri}. (1) \textbf{Killing Energy}. \(E_\text{kill}\) ensures accurate reproduction of isometric and rigid motions: applying rotation to a Genus-2 mesh from Thingi10K~\cite{Thingi10K} (ID: 83024), we accurately reconstruct the shape with \(E_\text{kill}\), whereas its removal introduces visible artifacts (Top). (2) \textbf{Dirichlet Energy}. \(E_\text{diri}\) enforces spatial smoothness, effectively capturing subtle deformations such as gentle wrist movements (Right). (3) \textbf{Consistent Speed}. \(E_\text{speed}\) produces temporally smooth and plausible motions even from just three input frames (Bottom). In contrast, DSR~\cite{sun2024dsr}, relying primarily on gradient-based optimization, struggles with near-isometric deformations.

\vspace{1mm}
\noindent\textbf{Runtime Efficiency.}  In \cref{tab:efficiency}, we evaluate fitting time (FitTime), interpolation time (InterTime), and velocity field sizes on sequence \texttt{rabbit7L6\_HiderotL} using an A100 80G GPU. At interpolation, we first extract a canonical mesh via marching cubes and deform it into 50 interpolated frames. NVFi has fewer parameters but incurs higher positional encoding overhead. DSR must extract meshes per frame, significantly increasing interpolation costs. Although DSR achieves the fastest fitting due to its simpler, frame-independent processing, our method produces superior results. Notably, our compact velocity field provides the fastest interpolation speed. Our fitting time includes velocity derivative computations, with 20GB GPU usage peak.
\begin{table}[!htb]
\vspace{-4mm}
\caption{\textbf{Efficiency.} \#velo indicates velocity field parameter count; FitTime is averaged per 100 iterations; InterTime includes canonical mesh extraction and deformation across the sequence.}
\label{tab:efficiency}
\vspace{-1mm}
\resizebox{\linewidth}{!}{
\begin{tabular}{c c c c c}
\toprule
Method & Category & \#velo $\downarrow$ & FitTime (s) $\downarrow$ & InterTime (s) $\downarrow$ \\
\midrule
DSR~\cite{sun2024dsr} & 4D Implicit & - & \cellcolor{LightCyan}\textbf{5.29} & 251 \\
OFlow~\cite{niemeyer2019occupancy} & ODE Flow & 3,291,395 & 83.4 & 2.48 + 40.3 \\
NDF~\cite{sun2022topology} & ODE Flow & 1,591,564 & 27.0 & 2.47 + 17.5 \\  
NVFi~\cite{li2024nvfi} & ODE Flow & \cellcolor{LightCyan}\textbf{70,534} & 36.2 & 2.47 + 21.6 \\
Ours & ODE Flow & 285,187 & 30.1 & \cellcolor{LightCyan}\textbf{1.30 + 15.6} \\
\bottomrule
\end{tabular}}
\vspace{-2mm}
\end{table}


\section{Conclusion}
\label{sec:conclusion}
We introduced a hybrid implicit-explicit framework for accurately reconstructing the geometry and motion of deforming objects. Our novel consolidation process shows promise for various applications, including motion capture and dynamic scene understanding. Looking ahead, integrating texture information and exploring alternative inputs (e.g., video) could further broaden our method's applicability.
\vspace{-4mm}
\paragraph{Limitations} Our method relies on explicit diffeomorphic flows and thus cannot handle topological changes (e.g., hole formation). It may also struggle to accurately interpolate motions that are excessively rapid between keyframes. Please refer to the supplementary material for more details.

\vspace{-4mm}
\paragraph{Acknowledgements} We thank Hakan Bilen and Oisin Mac Aodha for their help in preparing this manuscript.
{
    \small
    \bibliographystyle{ieeenat_fullname}
    \bibliography{main}
}


\maketitlesupplementary

\appendix
\setcounter{table}{0}
\renewcommand{\thetable}{A\arabic{table}}
\setcounter{figure}{0}
\renewcommand{\thefigure}{A\arabic{figure}}

\section{Organization}
This supplementary covers the diffeomorphic flow, velocity field analysis, articulated rigidity, dynamic consolidator architecture, multistage hyperparameters scheduling, complete implementation details, and additional experimental results. It also outlines additional robustness evaluations (missing regions, additional raw scans, sparse frames, sparse points, robustness to noise, topological artifacts, and normals), potential applications (arbitrary mesh discretization, consolidating textured scans, and dynamic texture generation), and limitations (topological changes, rapid motion transitions, and capturing finer geometric details).

\textbf{Note:} Figures, sections, and tables in the supplementary material are prefixed with a letter for distinction, while those without a prefix refer to content in the main paper.

\section{Diffeomorphic Flow}
We now present the core motivation of our methodology.

\paragraph{Temporal Coherence via Diffeomorphic Flows.} 4D Implicit methods~\cite{Eckstein2007,atzmon2021augmenting,stam2011velocity,tao2016near} represent shapes independently per frame and lose explicit pointwise correspondences over time, resulting in temporally incoherent interpolation, difficulty in handling missing regions (Fig.~9), and noises (Fig.~\ref{fig:uniform_noise}). To address this, we explicitly model deformation as a diffeomorphic flow via a continuous velocity field, and factor our representation into two parts:

\begin{enumerate}
    \item \textbf{Canonical Shape $g_c$}: An implicit spatial function $g_c(x):\mathbb{R}^3\rightarrow \mathbb{R}$ defined at canonical time $c=0.5$, implicitly representing the canonical shape $S_c=g_c^{-1}(0)$.
    \item \textbf{Velocity Field \( v \)}: A spatially and temporally varying velocity field \( v(x,t):\mathbb{R}^3\times\mathbb{R}\rightarrow\mathbb{R}^3 \) defining a continuous diffeomorphic flow.
\end{enumerate}

\paragraph{Diffeomorphic Flow via ODE Integration.}  
Given the velocity field \( v(x,t) \), we numerically integrate an ordinary differential equation (ODE) using standard solvers \cite{torchdiffeq} to obtain flow maps. Specifically, given a point \( x \) at time \(0\), its forward flow map \(\phi^{\rightarrow}\) to the position \( y \) at time \( t \) is defined as:
\begin{align*}
\begin{cases} 
\frac{\partial \phi^{\rightarrow}(x,\tau)}{\partial \tau} &= v(\phi^{\rightarrow}(x,\tau), \tau),\quad\tau \in [0,t)\\[5pt]
\phi^\rightarrow(x,0)&=x,\\[5pt]
\phi^\rightarrow(x,t)&=y.
\end{cases}
\end{align*}
Similarly, we define the backward flow map \(\phi^{\leftarrow}\). Practically, we denote the flow map from arbitrary time \(t\) to canonical time \(c=0.5\) as \(\phi(x,t\mapsto c)\), and its inverse for evaluation as \(\phi(x,c\mapsto t)\). Thus, our deforming shape representation is \( f(x,t) = g_c \circ \phi(x, t \mapsto c) \). This implicit-explicit decomposition offers several practical advantages:

\begin{itemize}
    \item Explicit velocity modeling induces a smooth diffeomorphic flow, naturally ensuring consistent pointwise correspondences without requiring separate forward-backward networks or cycle-consistency constraints (Fig.~\ref{fig:quad_remesh}).
    \item The canonical shape \( g_c \) aggregates geometric details from all frames, enabling robust reconstruction even with incomplete, sparse, or noisy data (Fig.~9, Fig.~\ref{fig:uniform_noise}, Fig.~\ref{fig:spar_pts}).
    \item Using an implicit canonical shape flexibly allows the reconstruction of arbitrary topologies.
    \item Decoupling shape and motion representation, where the canonical shape captures fine geometric details and the velocity field encourages smooth deformation,
\end{itemize}

\section{Velocity Field Analysis}
\begin{figure}[!htb]
    \centering
    \includegraphics[width=\linewidth]{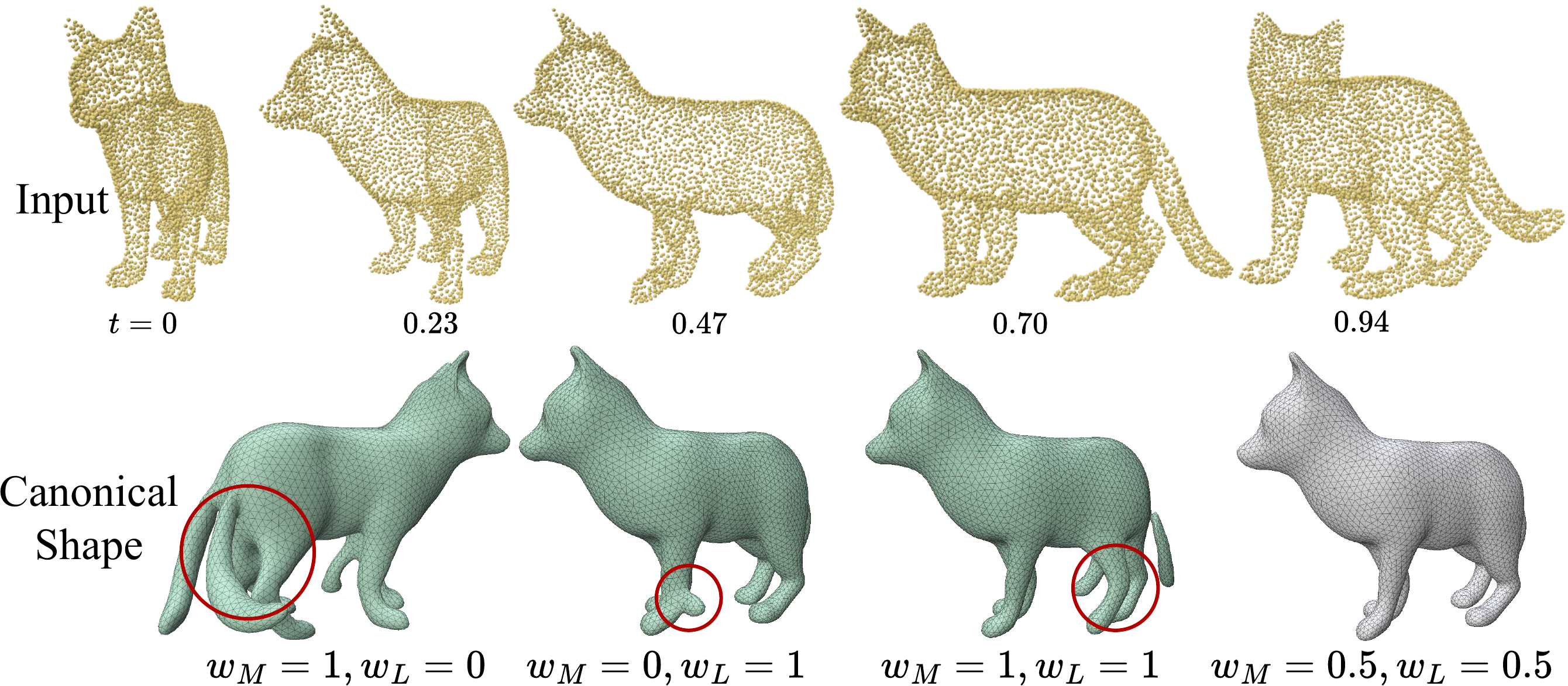}
    \vspace{-6mm}
    \caption{\textbf{Additional Velocity Ablation.} Given sequential point cloud inputs, different canonical shapes can be reconstructed when setting different $w_M$ and $w_L$ values. An average weighting scheme works best, and we adopt it throughout our experiments. 
    }
    \label{fig:velocity_abl}
\end{figure}

We bias our velocity representation towards a mixture of low- and medium-frequency velocities. With this, we mitigate kinks and abrupt changes in the deformation. In addition to the Fourier encoding in space-time coordinates in Sec. 5.1, we use a mixture of two MLP layers to represent the velocity field itself (see Fig. 6). Consider two fully connected single-layer networks 
 $\zeta_A:\mathbb{R}^d\rightarrow \mathbb{R}^h$ and $\zeta_B:\mathbb{R}^h\rightarrow\mathbb{R}^h$ and   with $h=512$. 
A SoftPlus~\cite{dugas2000incorporating} function activates these MLP layers. The purpose of this composition is to use the smooth attenuation bias of neural networks, where $\zeta_B \circ \zeta_A$ serves as a low-frequency component; using it alone (i.e., $w_M=0$) would result in the loss of small details in the deformation. Using $\zeta_A$ alone (i.e., $w_L=0$), as the medium-frequency component, results in failure to capture global motion. In Fig.~\ref{fig:velocity_abl}, we ablate these weights and demonstrate that the average weighting (i.e., $w_L=w_M=0.5$) is a successful choice, even when compared to the conventional skip connection $w_L=w_M=1$. 

\section{Articulated Rigidity Details} 
\begin{figure}[!htb]
    \centering
    \includegraphics[width=\linewidth]{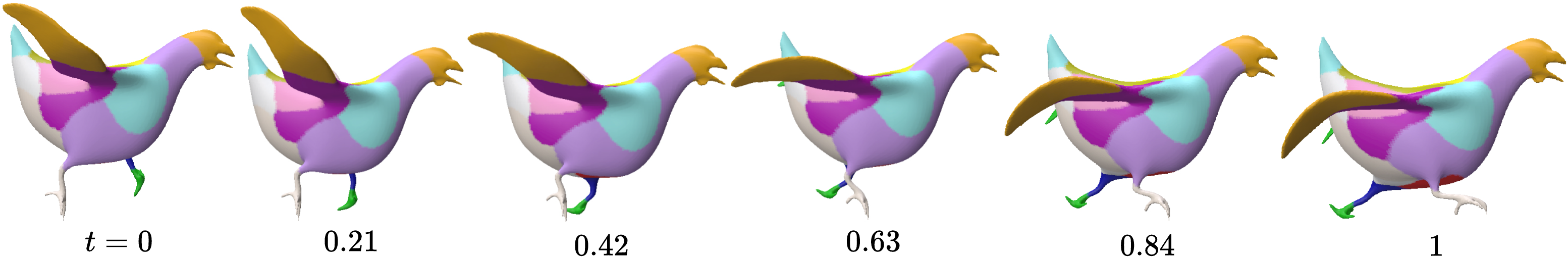}
    \vspace{-6mm}
    \caption{\textbf{Motion Segmentation.}  The motion-segmentation network learns the articulation of the object and segments the (softly) piecewise-rigidly moving parts (different colors). 
    }
    \label{fig:part_seg}
\end{figure}

Following \cite{atzmon2021augmenting, zhang2023self}, physical objects, especially live creatures, tend to move in (softly) rigid semantic parts, such as the limbs between the joints. These kinematic units constitute a quasi-articulated system. For this, we adapt a motion-segmentation module from~\cite{zhang2023self}. For a prescribed number of segments $H=20$ (See the Supplementary in \citet{zhang2023self} for the ablations), we learn a motion-segmentation network that is implemented as a neural field $\zeta_H:\mathbb{R}^3 \rightarrow [0,1]^H$, that for each location $x$ in the space of the canonical time $c$ outputs a probability $\zeta_{H,h}(x)$ of the point $x$ belonging to segment $h$ (as a partition of unity: $\sum_h{\zeta_{H,h}(x)}=1$). 
We add a loss term that regularizes the per-part rigidity, by computing a pair of the rotation matrix and translation vector $\left\{R_{h,i} \in SO(3),\ \tau_{h,i} \in \mathbb{R}^3\right\}$ per each segment $h$ and time frame $i$, where we seek that the flow from $t_i$ to $c$ is as-(part-wise)-rigid-as-possible \cite{sorkine2007rigid}, weighted by the probability of $h$:
\begin{equation}
\begin{split}
   E_\text{rigid}(i) =   \mathbb{E}_{x_j\in \mathcal{P}_i} \sum_{h \in [1,H]} \zeta_{H,h}(\phi(x_j,t_i\mapsto c))\\
   \cdot
   \norm{(R_{h,i}x_j+\tau_{h,i})-\phi(x_j,t_i\mapsto c)}^2,
\end{split}
\end{equation}
In each iteration, we compute the rigid transformation $\left\{R_{h,i},\tau_{h,i}\right\}$ in closed-form by SVD of the correlation matrix between the segment at time $t$ and that of time $c$ (the classical ARAP local step). As a by-product of this fitting, we get a segmentation of the moving object, by considering the highest probability of each point (Fig.~\ref{fig:part_seg}).

\section{Experiments}
\subsection{Experimental Details}
\begin{figure}[!htb]
\centering
\includegraphics[width=\linewidth]{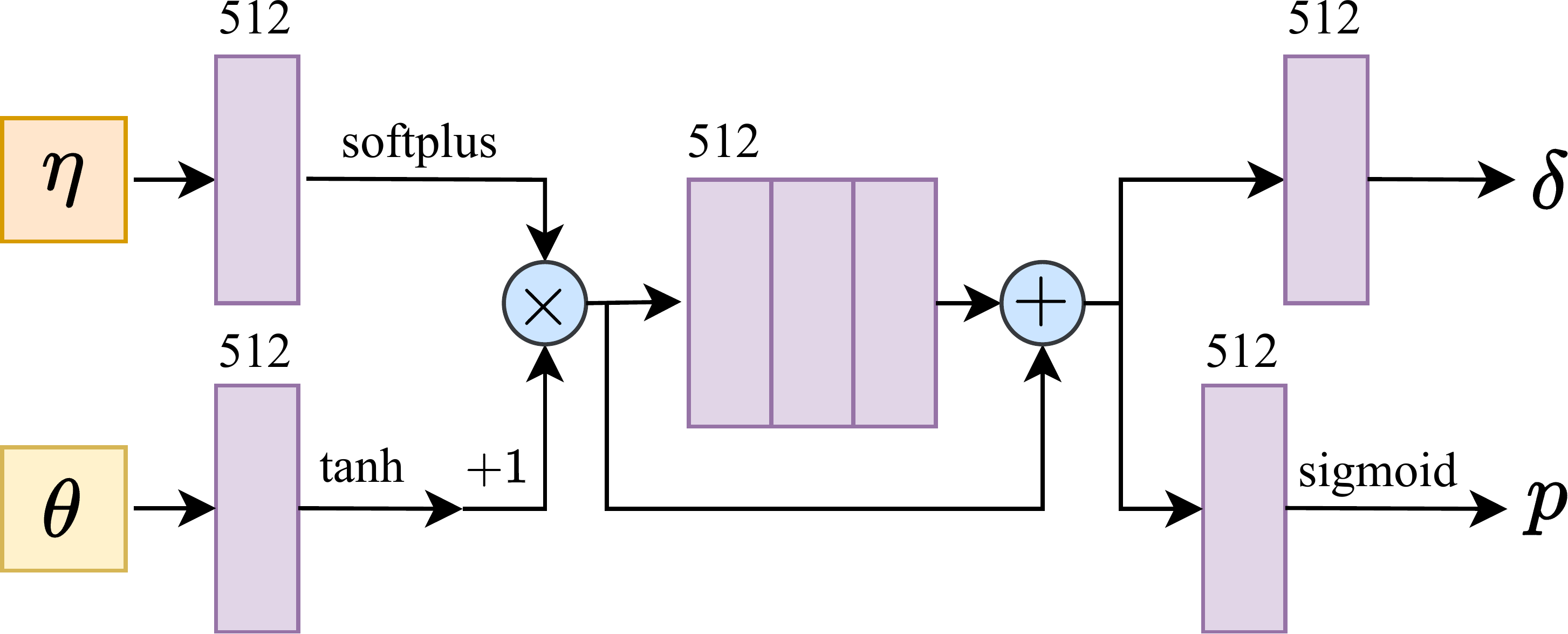}
\caption{\textbf{Dynamic consolidator architecture.}}
\label{fig:consolidator_structure}
\end{figure}

\paragraph{Dynamic Consolidator.}
Our dynamic consolidator (Sec. 4) takes as input a spacetime coordinate \(\eta = (x,t)\) and an optimizable latent code \(\theta\), and outputs the perturbation \(\delta\) and confidence score \(p\). As illustrated in \cref{fig:consolidator_structure}, it consists of MLP layers with 512 neurons per layer.

\paragraph{Multistage Hyperparameters Scheduling.} 
Our experiments are run for 15000 full learning iterations. We implement a scheduling strategy for the velocity-network regularization weights $\lambda_\text{kill}$, $\lambda_\text{diri}$, and $\lambda_\text{speed}$ as follows: they are initially set to 0 for the first $7000$ iterations (during which we do not compute the unused associated derivatives); by doing so, the velocity field is only guided by $\widehat{E}_{\text{fit}}$, $E_{\text{eik}}$ and $E_{\text{rigid}}$. We subsequently increase them as follows: from
$\lambda_\text{kill}=\expnumber{5}{-5}$, $\lambda_\text{dir}=\expnumber{8}{-5}$, and  $\lambda_\text{speed}=\expnumber{1}{-3}$ to $\expnumber{1.5}{-3}$, $\expnumber{3.8}{-3}$, and $\expnumber{2}{-2}$ respectively and uniformly over the next $5000$ iterations. We then maintain them at $\expnumber{5}{-4}$, $\expnumber{6.5}{-4}$, and $\expnumber{5}{-3}$ respectively for the final $3000$ iterations. We set the fixed $\lambda_\text{rigid} = \expnumber{1}{3}$, based on \cite{zhang2023self}, and also set $\lambda_\text{eik}=0.1$. 
Our loss coefficients $\lambda_\text{mag}$, $\lambda_\text{var}$, and $\lambda_\text{log}$ for the dynamic consolidator, unless otherwise specified (e.g., Fig.~\ref{fig:topology}), are initially set to $\lambda_\text{mag}=\lambda_\text{var}=0.01$, and $\lambda_\text{log}=0.10$ for the first $3000$ iterations. We then increase them to $0.4$, $30$, and $2.0$ respectively and uniformly over the next $9000$ iterations, maintaining these final values for the last $3000$ iterations. 

\paragraph{Code and Hardware.} We run all our experiments on a single NVIDIA A100 80GB GPU. Our code is based on PyTorch~\cite{paszke2019pytorch} and uses the torchdiff~\cite{torchdiffeq} package to implement the ODE solver. We use Polyscope~\cite{polyscope} for visualization.

\paragraph{ODE Solver.} To integrate the flow $\phi$ from the velocity $v$, we use the Dormand–Prince method `dopri5'~\cite{dormand1980family}, setting relative and absolute error tolerances to $\expnumber{1}{-3}$ and $\expnumber{1}{-5}$, respectively.
\subsection{Additional Experiment Results}

\begin{figure*}[t]
\centering
\includegraphics[width=\linewidth]{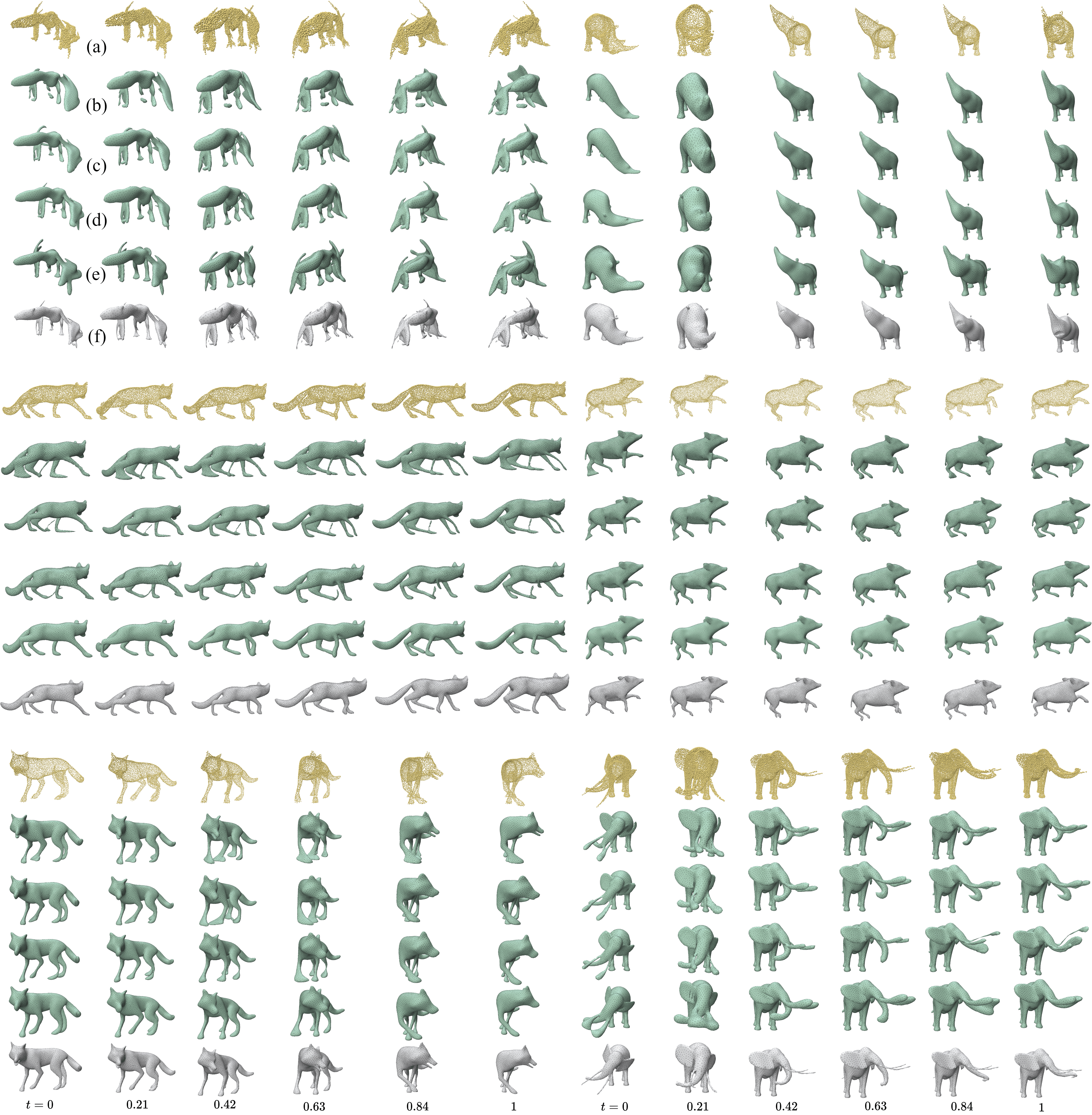}
\caption{\textbf{Additional Qualitative Comparisons.} Synthetic animal motions from DeformingThings4D~\cite{li20214dcomplete}: (a) Point clouds, (b) OFlow~\cite{niemeyer2019occupancy} + NVFi velocity~\cite{li2024nvfi}, (c) NDF~\cite{sun2022topology}, (d) OFlow~\cite{niemeyer2019occupancy}, (e) DSR~\cite{sun2024dsr}, and (f) Ours. Our method reconstructs natural temporal deformations, preserving geometric details without oversmoothing or introducing topological artifacts.}
\label{fig:qua_animal}
\end{figure*}

\paragraph{Additional Qualitative Comparisons.} 
We provide additional qualitative results for 4D interpolation of synthetic animals from DeformingThings4D in \cref{fig:qua_animal}.

\subsection{Additional Robustness Evaluations}
\paragraph{Missing Regions.}
In addition to qualitative evaluations in Fig.~9, we quantitatively compare our approach with DSR~\cite{sun2024dsr} on partially missing inputs from the animal dataset. Metrics in Tab.~\ref{tab:miss} confirm our method’s superior performance.
\begin{table}[!htb]
  \fontsize{15}{20}\selectfont
  \centering
    \caption{\textbf{Quantitative Comparison: Inputs with Missing Regions.} Evaluation corresponds to Fig.~9 (animal dataset).}
    \label{tab:miss}
  \resizebox{\linewidth}{!}{
  \begin{tabular}{c c c c c c c c}\hline
    \toprule
     \multicolumn{1}{c}{\multirow{2}{*}{}}&\multirow{2}{*}{Sequence Name} &\multirow{2}{*}{Method}& \multicolumn{2}{c}{IoU(\%)} & \multicolumn{2}{c}{CD($\times 10^{-5}$)}\\
    \cmidrule(lr){4-5}
    \cmidrule(lr){6-7}
    \multicolumn{1}{c}{}& &&Mean$\uparrow$ & Min$\uparrow$ & Mean $\downarrow$ & Max$\downarrow$\\
    \midrule
    &\multirow{2}{*}{\texttt{deerFEL\_WalkhuntedRM}}&DSR\cite{sun2024dsr} & 80.01 & 74.95 & 65.72 & 168.0\\
    &&\cellcolor{LightCyan}\textbf{Ours}&\cellcolor{LightCyan} \textbf{89.06} & \cellcolor{LightCyan}\textbf{82.19} &\cellcolor{LightCyan} \textbf{6.819} & \cellcolor{LightCyan}\textbf{19.39} \\
    \midrule
    &\multirow{2}{*}{\texttt{bear3EP\_WalkrightRM}}&DSR\cite{sun2024dsr} & 83.11 & 78.71 & 118.0 & 441.8\\
    &&\cellcolor{LightCyan}\textbf{Ours} & \cellcolor{LightCyan}\textbf{89.38} &\cellcolor{LightCyan}\textbf{86.21}  &\cellcolor{LightCyan}\textbf{36.80} & \cellcolor{LightCyan}\textbf{90.90} \\ 
    \bottomrule
  \end{tabular}
  }
\end{table}

\begin{figure}[!htb]
\centering
\includegraphics[width=\linewidth]{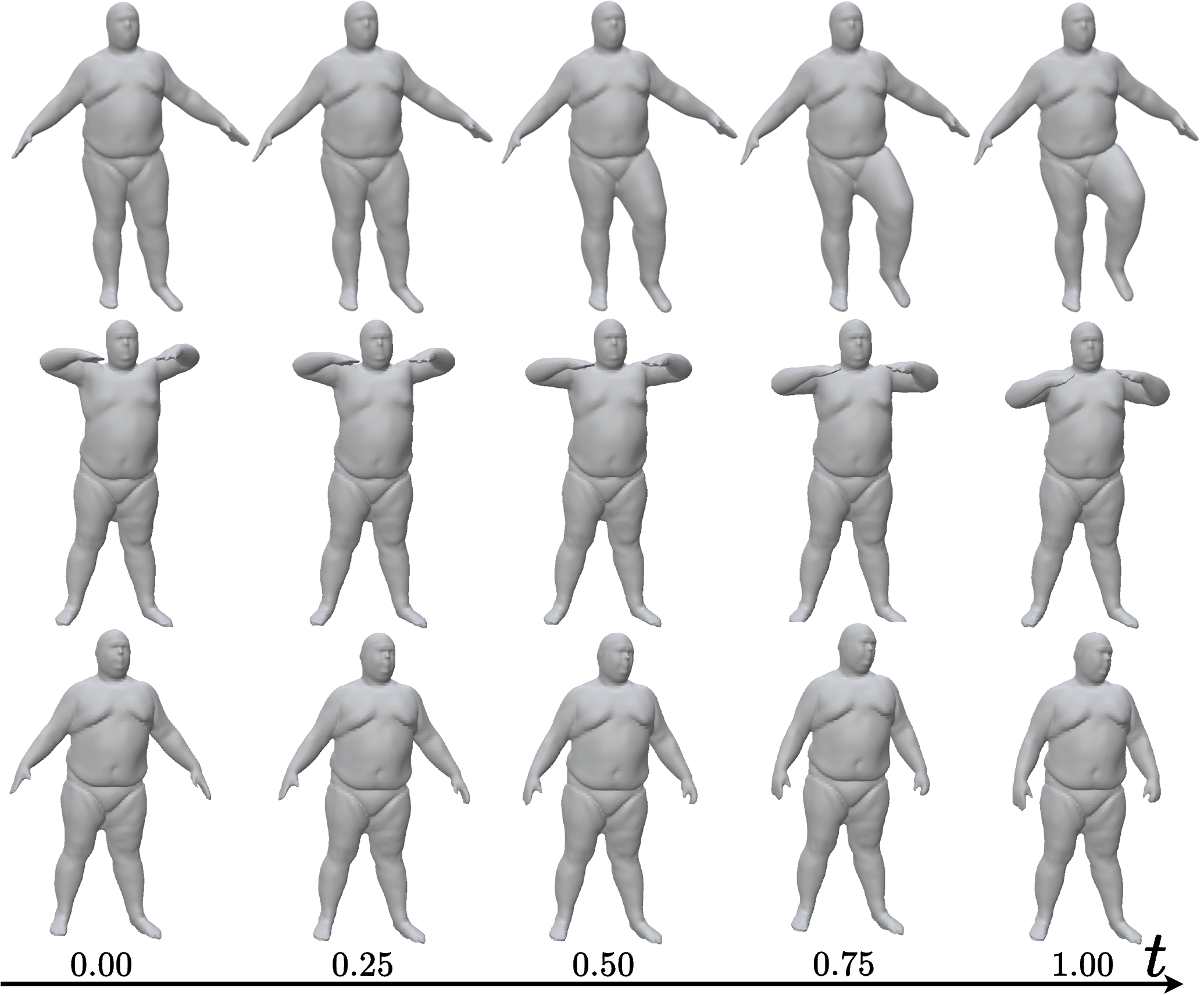}
\caption{\textbf{Additional Raw Scans.} From top to bottom: \texttt{one\_leg\_jump}, \texttt{chicken\_wings}, \texttt{light\_hopping\_stiff} from DFAUST (Subject ID = 50002).}
\label{fig:additional_raw}
\end{figure}

\paragraph{Addtional Raw Scans.} 
Further interpolation results on raw DFAUST scans using our approach are shown in \cref{fig:additional_raw}.

\begin{figure}[!htb]
\centering
\includegraphics[width=\linewidth]{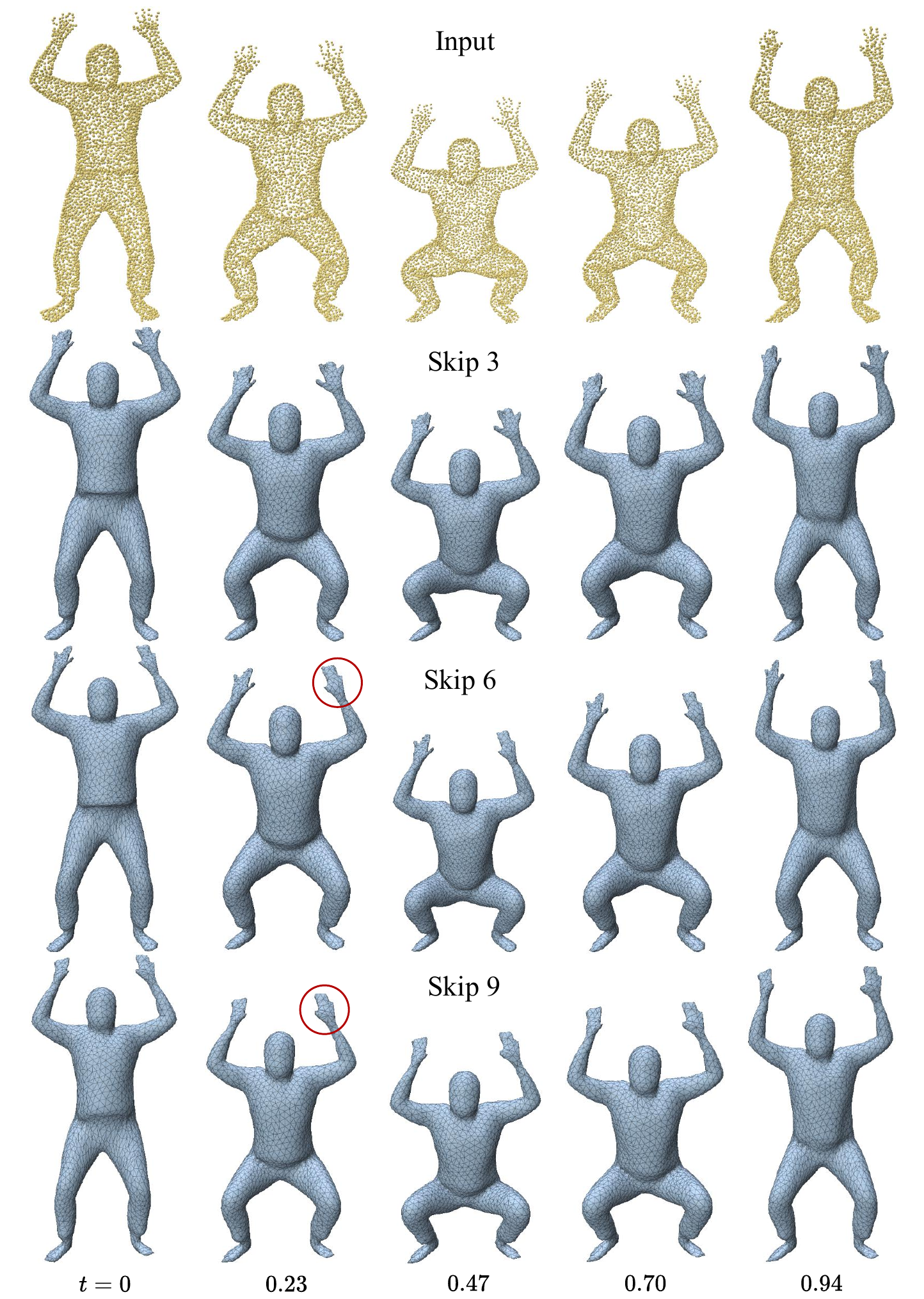}
\vspace{-7mm}
\caption{\textbf{Robustness to Frame Sparsity.} Interpolation results at identical time frames using varying input sparsity. "Skip" denotes temporal sampling intervals from the original sequence (e.g., a skip of 3 means using frames $1,4,7,\dots$). Our method accurately reconstructs motions despite substantial frame skipping.}
\label{fig:regular_sparse}
\end{figure}

\begin{figure}[!htb]
\centering
\includegraphics[width=\linewidth]{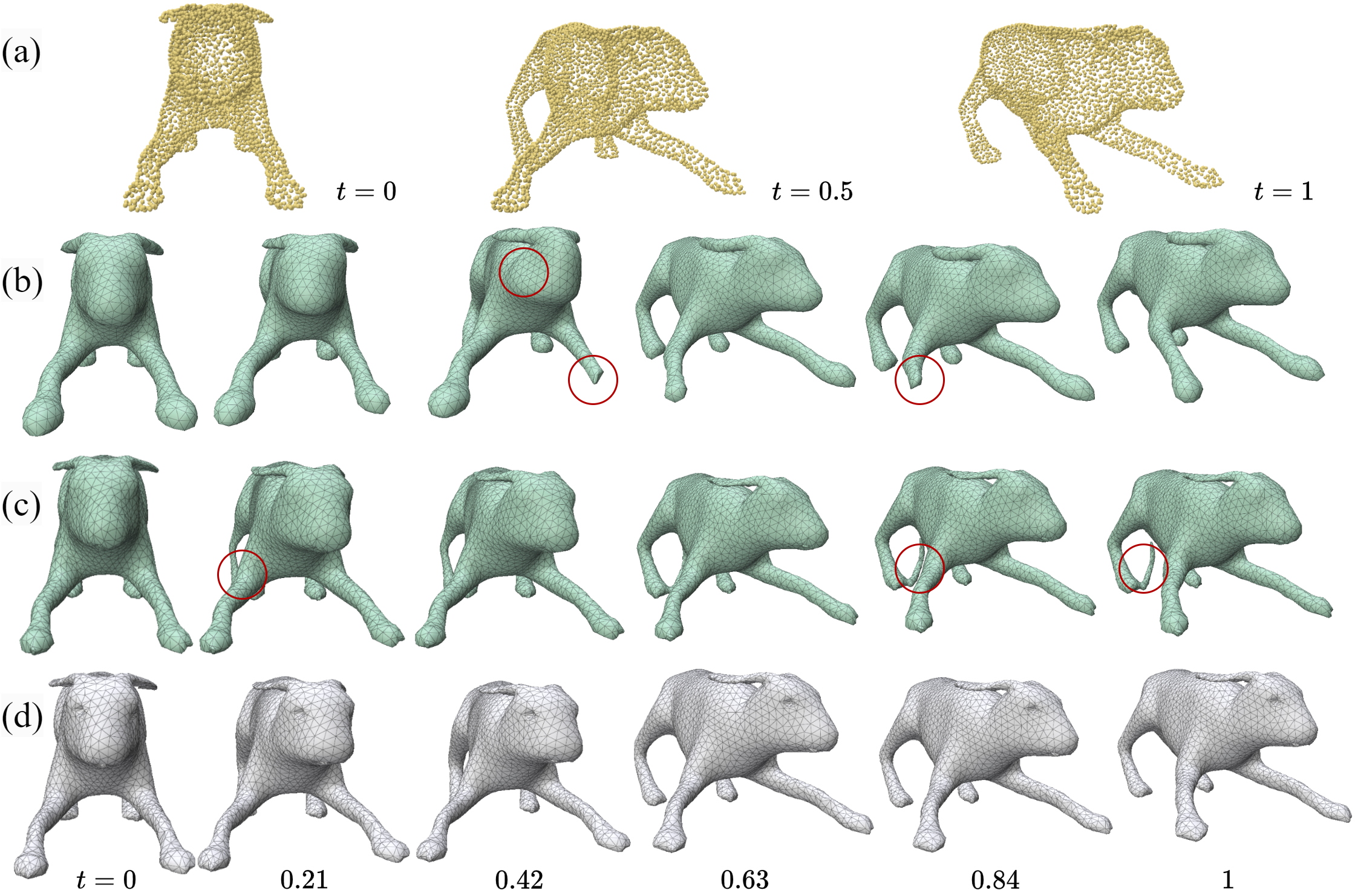}
\caption{\textbf{Comparison under Extremely Sparse Input.} (a) Three-frame Input, (b) DSR, (c) Ours without speed-consistency loss ($E_\text{speed}$), (d) Ours (full). Our full approach achieves better interpolation quality and fewer visual artifacts (circled).}
\label{fig:sparse}
\end{figure}

\paragraph{Sparse Frames.}
In Fig.~\ref{fig:regular_sparse}, we show that our method can smoothly reconstruct motions even from very sparse input frames. As illustrated in Fig.~\ref{fig:sparse}, the competing method DSR~\cite{sun2024dsr} struggles to maintain near-isometric deformations, likely because it follows the network gradient rather than physically plausible motions. Furthermore, this comparison highlights the importance of our proposed speed-consistency loss ($E_\text{speed}$), which helps preserve small-scale structures (e.g., rabbit eyes in Fig.~\ref{fig:sparse}) and reduces interpolation artifacts when the input frames are limited.

\begin{figure}[!htb]
\centering
\includegraphics[width=\linewidth]{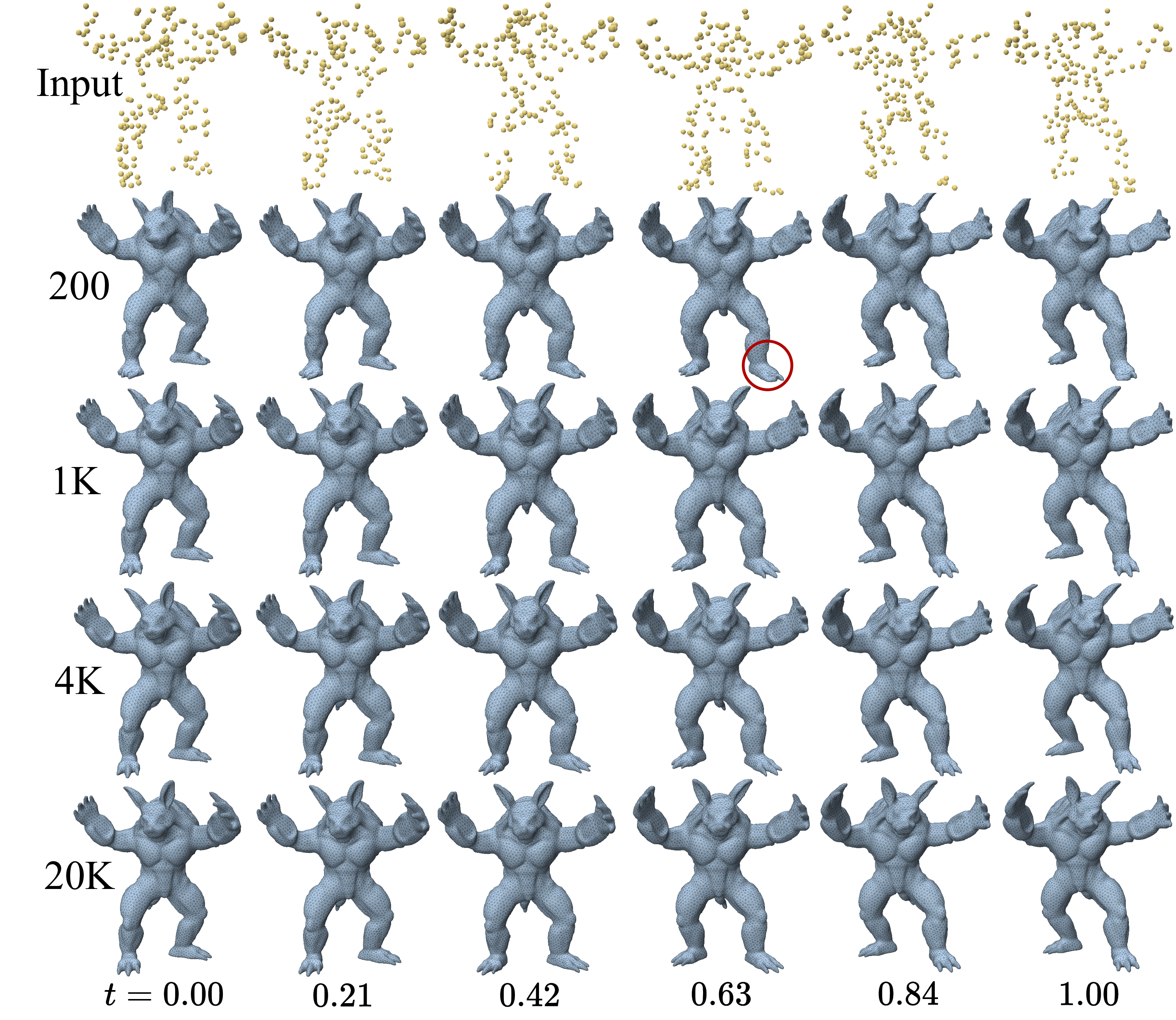}
\caption{\textbf{Robustness to Sparse Point Input.} Using as few as 200 points per frame maintains shape reconstruction, with gradual detail loss. Results using 4K and 20K points per frame are nearly indistinguishable.}
\label{fig:spar_pts}
\end{figure}

\paragraph{Sparse Points.}
Fig.~\ref{fig:spar_pts} tests our robustness to input point sparsity with a fixed sequence of 14 frames. Remarkably, even with only 200 points per frame (a total of $14\times200$ points consolidated at canonical time), we accurately reconstruct overall geometry and deformation, though fine details are naturally reduced. Increasing the number of points per frame to 20K does not further improve reconstruction quality, indicating a possible saturation due to the underlying SIREN representation. Unlike implicit approaches that independently handle each frame without aggregation, our flow-based method consolidates points from all frames into a canonical shape, significantly boosting robustness.

\begin{figure}[!htb]
\centering
\includegraphics[width=0.95\linewidth]{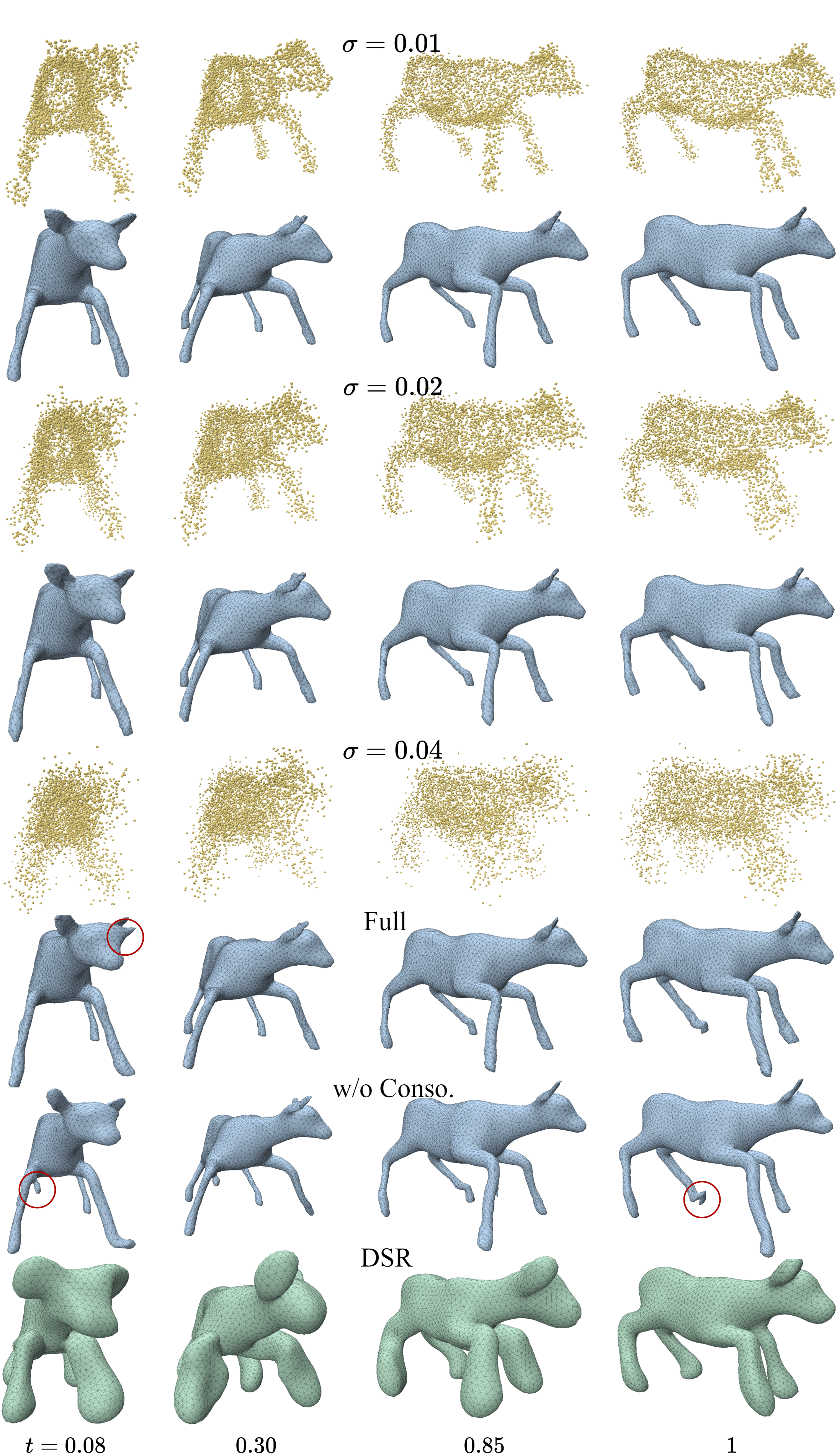}
\caption{\textbf{Robustness to Gaussian Noise.} Our method (blue) maintains robustness across various noise levels. At high noise ($\sigma=0.04$), removing the consolidator leads to visible artifacts. DSR (green) is particularly prone to noise, as highlighted by circled artifacts.}
\label{fig:uniform_noise}
\end{figure}

\paragraph{Robustness to Noise.}
We evaluate our framework under noisy conditions in Fig.~\ref{fig:uniform_noise}. Gaussian noise is added to 4 randomly selected frames within a 13-frame input sequence. Our method, especially the consolidator module, effectively filters noise-induced artifacts compared to DSR~\cite{sun2024dsr}, which produces overly smoothed geometry due to global sequence fitting.

\begin{figure}[!htb]
\centering
\includegraphics[width=\linewidth]{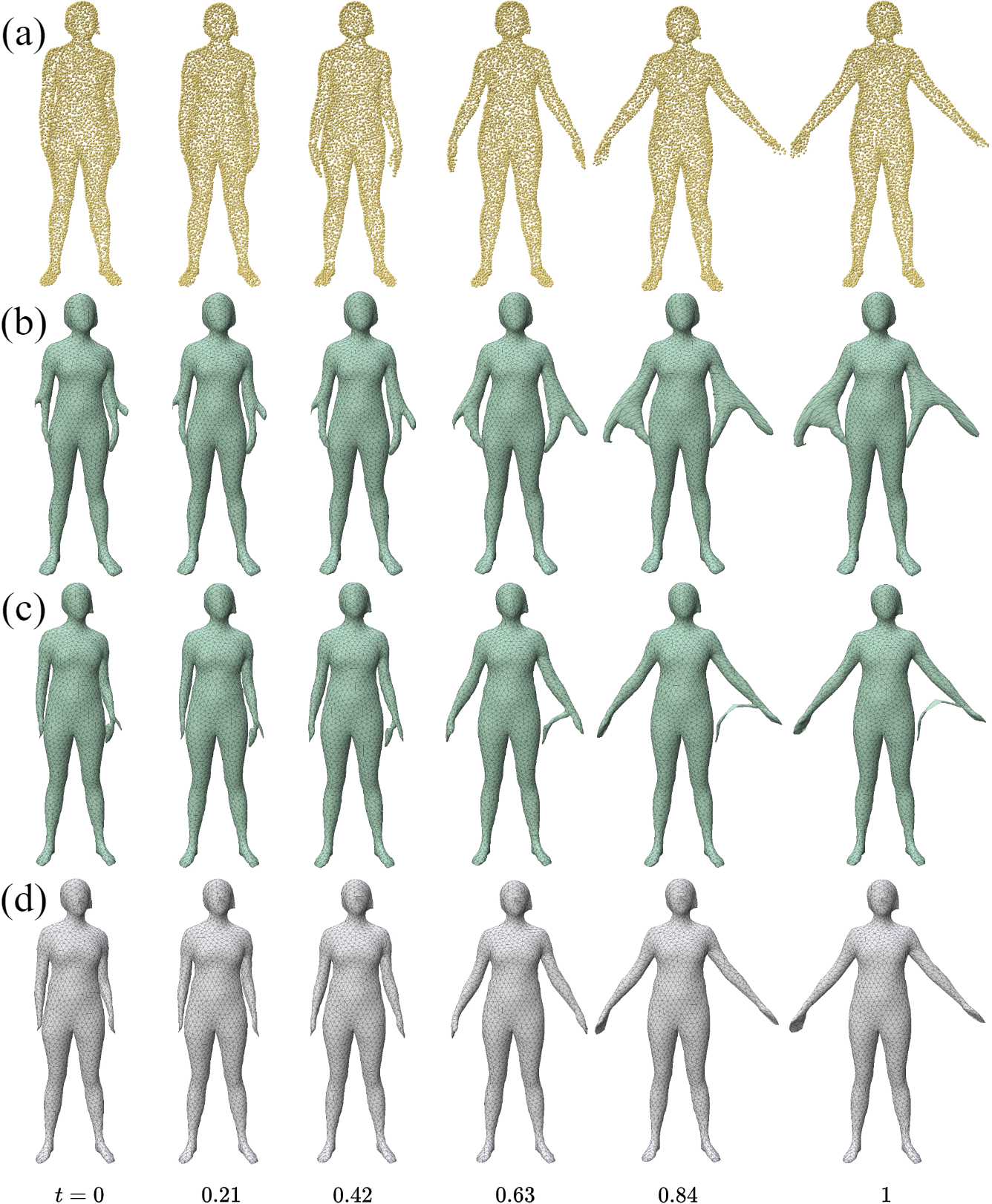}
\caption{\textbf{Robustness to Topological Errors.} (a) Input sequence; (b) OFlow~\cite{niemeyer2019occupancy}+NVFi velocity~\cite{li2024nvfi}; (c) Ours w/o consolidator; (d) Ours (full), which accurately reconstructs motion with minimal artifacts.}
\label{fig:topology}
\end{figure}

\paragraph{Topological Artifacts.}
Flow-based methods are sensitive to incorrect canonical shape inference, motivating our consolidator module. In Fig.~\ref{fig:topology}, we illustrate topology inference errors, such as mistakenly merging the hand into the torso. Compared to OFlow~\cite{niemeyer2019occupancy}+NVFi~\cite{li2024nvfi}, which infers incorrect topology, and our framework without the consolidator (correct topology but geometric artifacts), our full approach accurately preserves topology and geometry throughout deformation. Here we adjust the default weight $w_\text{log}$ to $0.005$ to slightly reduce early reliance on raw points.

\begin{figure}[!htb]
\centering
\includegraphics[width=\linewidth]{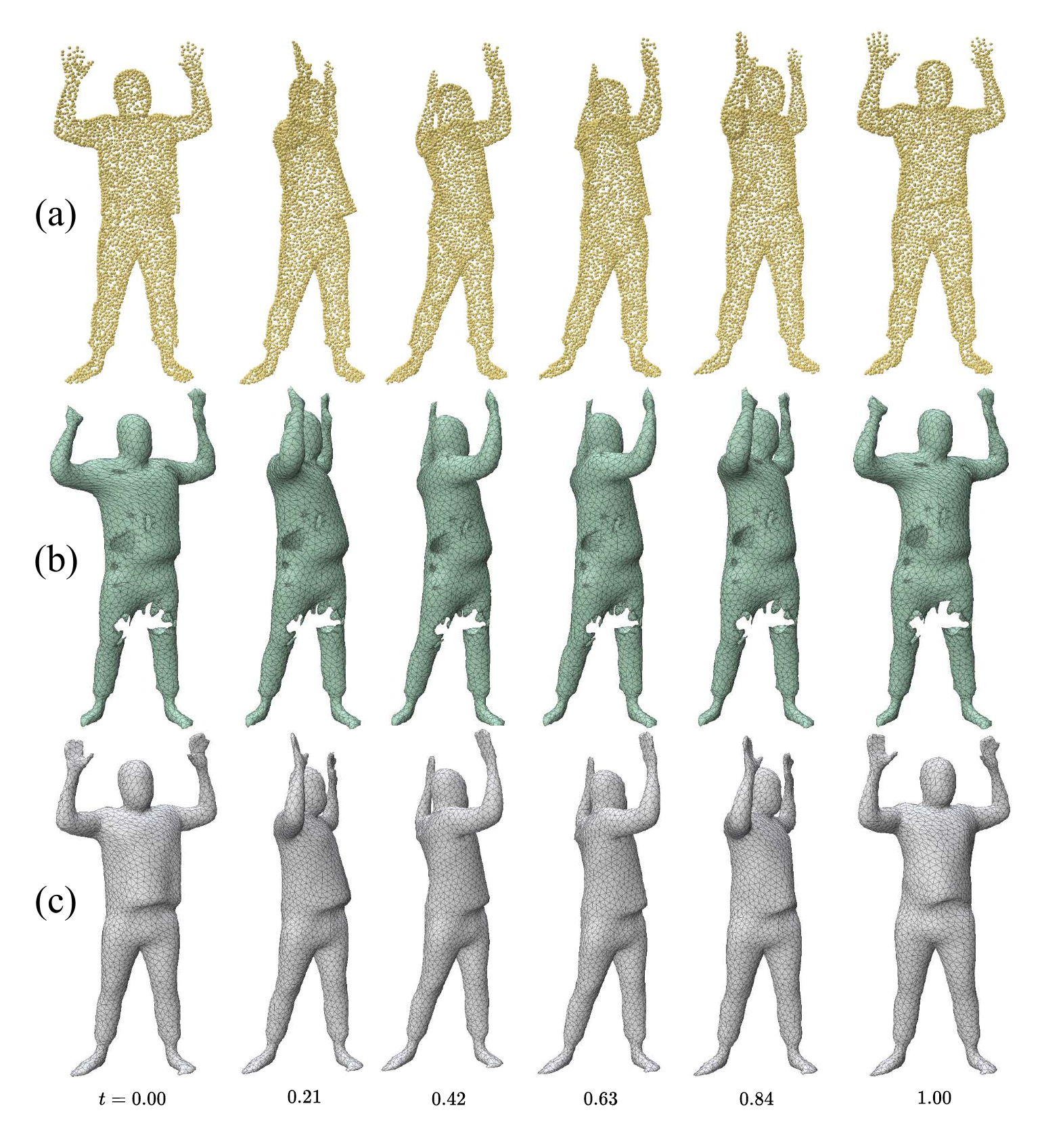}
\caption{\textbf{Effect of Missing Normals.} (a) Input; (b) Ours without normals; (c) Ours (full). Normals are crucial for accurate geometry reconstruction.}
\label{fig:wo_normals}
\end{figure}

\paragraph{Normals.}
For input point clouds without provided normals, we substitute $E_\text{fit}$ with the SALD loss term~\cite{atzmon2020sal}, which fits the unsigned distance function. Fig.~\ref{fig:wo_normals} reveals that this change significantly compromises geometric reconstruction, demonstrating the importance of normals in our formulation. Future research will explore alternative loss functions to address the absence of normal data.

\subsection{Applications}
\begin{figure}[!htb]
\centering
\includegraphics[width=\linewidth]{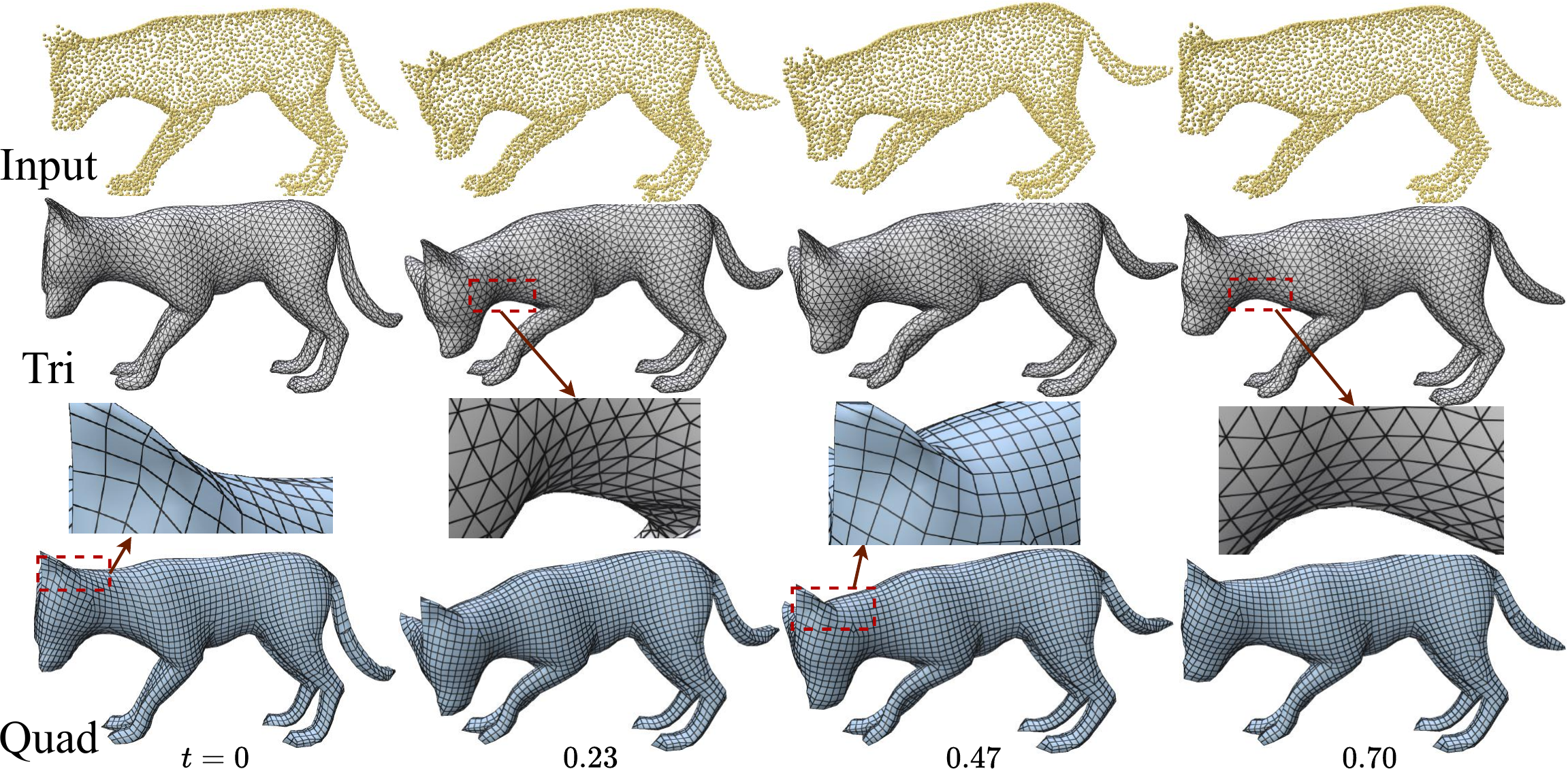}
\caption{\textbf{Mesh Representation Flexibility.} Our implicit canonical representation allows natural deformation on both triangular and quadrilateral meshes, without introducing noticeable tangential distortions.}
\label{fig:quad_remesh}
\end{figure}

\begin{figure}[!htb]
\centering
\includegraphics[width=\linewidth]{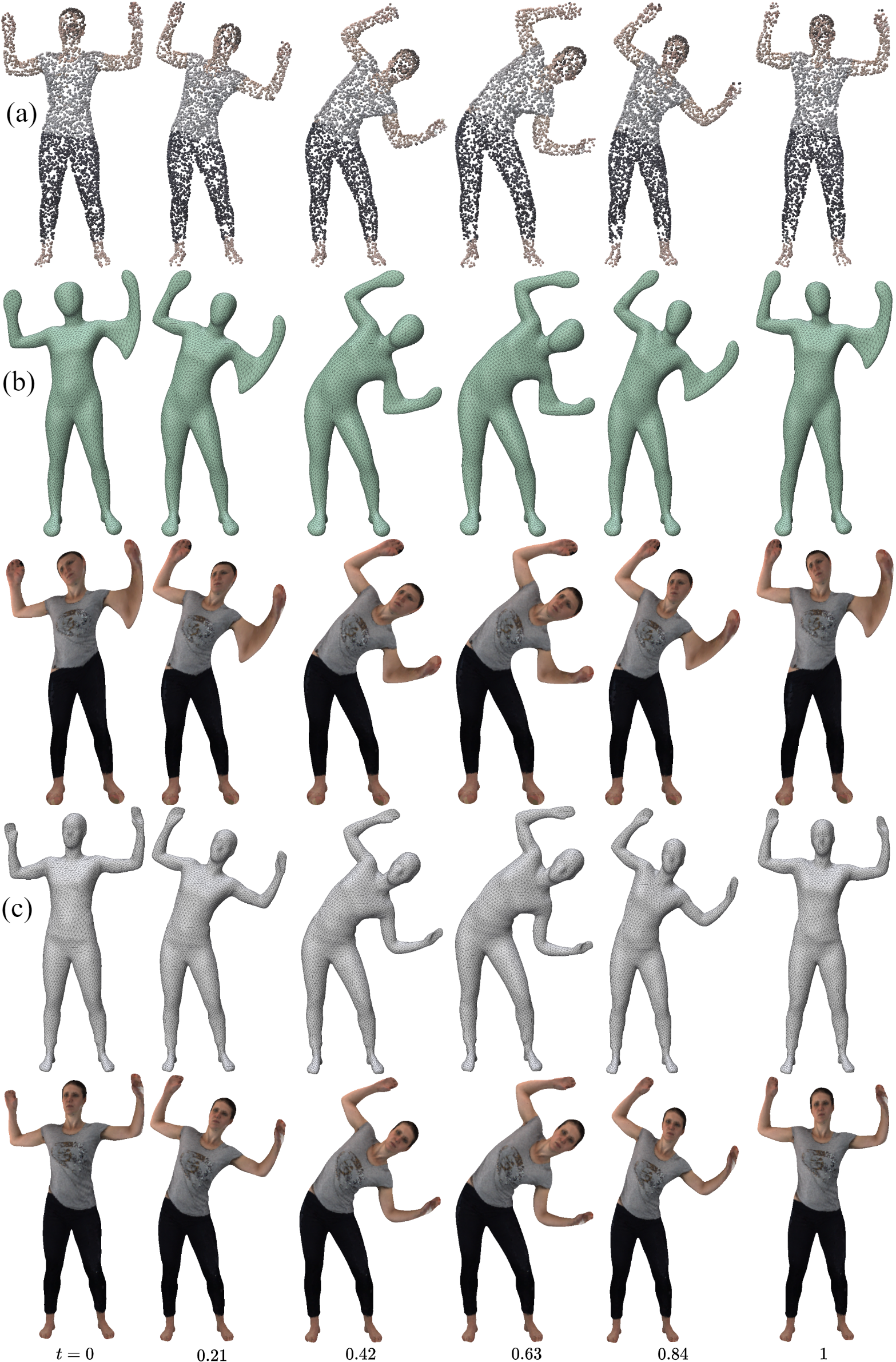}
\caption{\textbf{Consolidating Textured Scans.} (a) Input point clouds (4K points per frame) sampled from raw textured scans. (b) Result from NVFi. (c) Our result, demonstrating accurate geometry and texture preservation.}
\label{fig:real_scan}
\end{figure}

\begin{figure}[!htb]
\centering
\includegraphics[width=\linewidth]{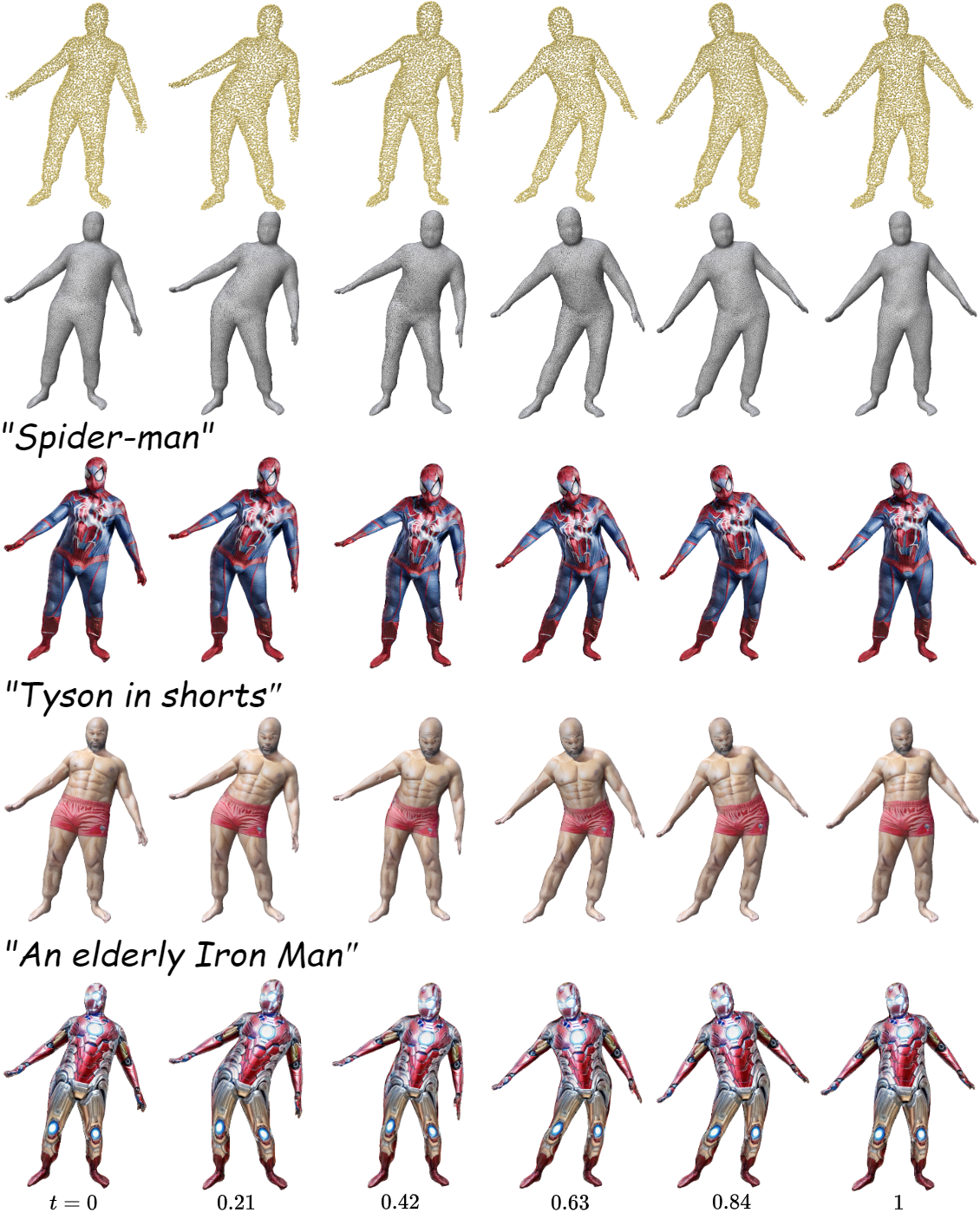}
\caption{\textbf{Dynamic Texture Generation.} We generate textures on the canonical frame (left) and consistently advect them over time, achieving temporally coherent textured animations.}
\label{fig:texture_gen}
\end{figure}

\paragraph{Arbitrary Mesh Discretization.}
Our method outputs an implicit canonical representation $g_c$, allowing flexibility in mesh discretization. Once a desired mesh (representing $S_c$) is extracted, vertex positions are directly updated via the learned velocity field to generate deformations. In Fig.~\ref{fig:quad_remesh}, we exemplify this versatility using both triangular and quadrilateral meshes, demonstrating intuitive deformations without significant distortion. The quadrilateral mesh is obtained using Instant Meshes~\cite{jakob2015instant}.
\paragraph{Consolidating Textured Scans.}
Our explicit modeling of flow naturally supports the temporal propagation of vertex attributes such as textures. In Fig.~\ref{fig:real_scan}, we demonstrate this capability using CAPE~\cite{pons2017clothcap} raw scans. We sample 4K textured points from input scans, associate these textures directly with vertices on the canonical shape, and consistently advect textures through the velocity field to other time frames. Compared to NVFi, our approach yields more accurate and visually coherent textured mesh sequences.
\paragraph{Dynamic Texture Generation.}
Our approach readily integrates with mesh texturing workflows. We illustrate in Fig.~\ref{fig:texture_gen} how we first apply Meshy~\cite{Meshai} to generate textures on our canonical shape based on textual prompts, then propagate these textures dynamically using our learned velocity field.

\subsection{Limitations}
Despite the promising performance of our method, several challenges remain open for future exploration:

\paragraph{Topological Changes.}
Our approach assumes consistent topology under diffeomorphic deformation, making it well-suited to applications such as motion capture or studying individual object deformation. However, unlike purely implicit methods (e.g., DSR), which inherently allow topological changes such as splitting or merging, our method cannot model scenarios like protein recombination or molecular bond-breaking (see supplementary video).

\begin{figure}[!htb]
\centering
\includegraphics[width=\linewidth]{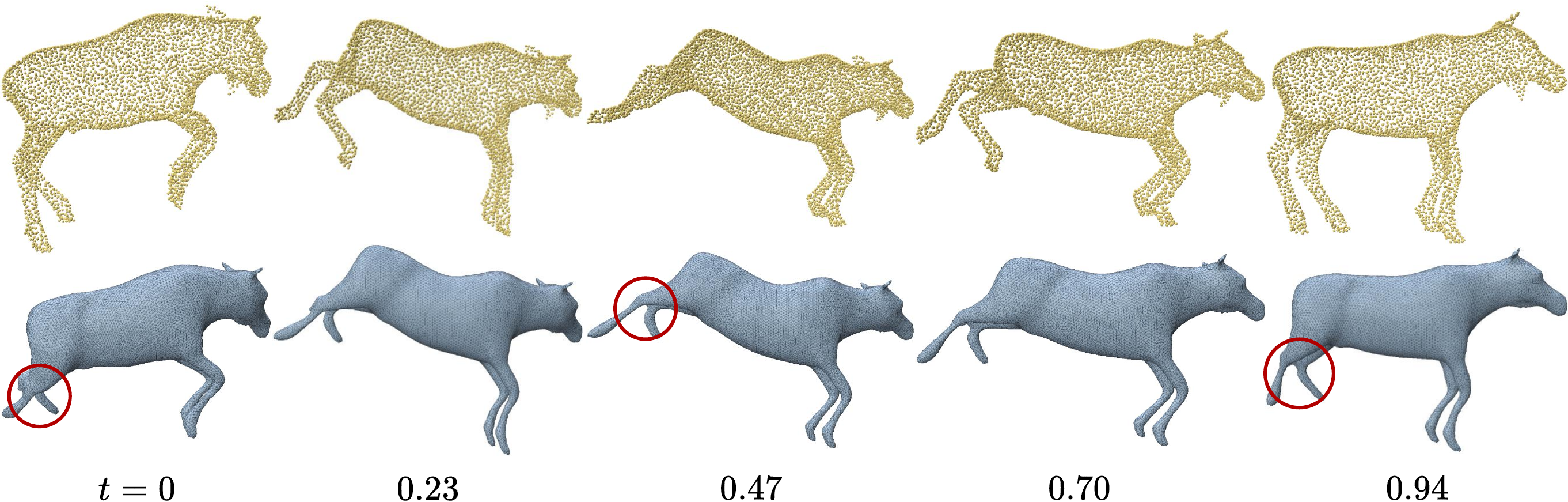}
\caption{\textbf{Failure with Rapid Motions.} Fast movements between frames can cause geometric artifacts (circled) due to insufficient temporal resolution.}
\label{fig:fail_case}
\end{figure}

\paragraph{Rapid Motion Transitions.}
Our approach may introduce geometric artifacts when motions between consecutive frames exhibit large variations, even if the deformation remains nearly isometric and topology consistent. Fig.~\ref{fig:fail_case} illustrates such a scenario. Although fine-tuning regularization terms might reduce this issue, we leave this investigation as future work

\paragraph{Capturing Finer Geometric Details.}
While our method captures geometric details better than many existing approaches, reconstructing extremely fine-scale features remains challenging (see Fig.~\ref{fig:spar_pts}). Improving high-frequency detail reconstruction continues to be an important open problem in geometry processing.

\subsection{Questions and Answers}

\begin{figure}[!htb] \centering 
\includegraphics[width=1.\linewidth]{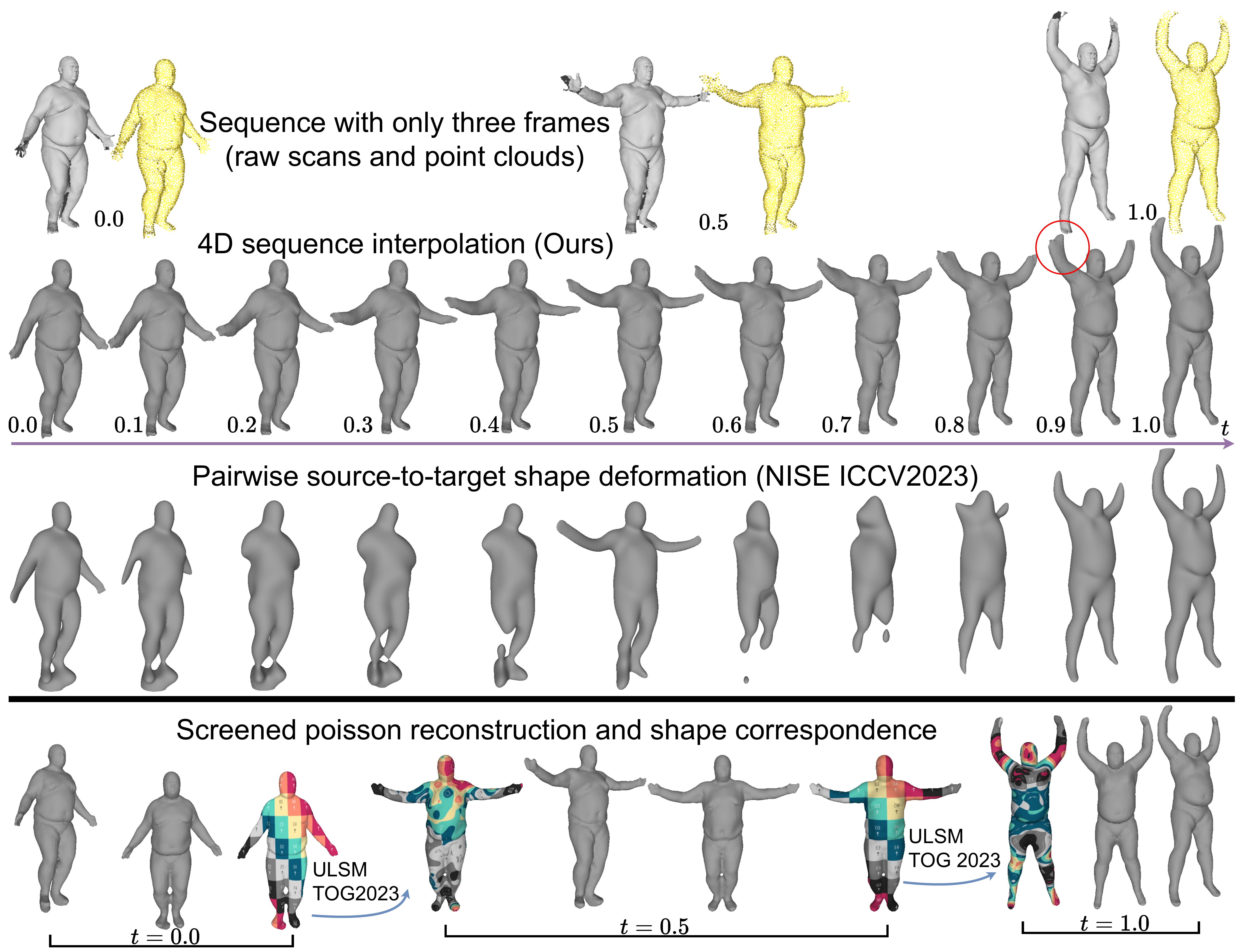}
\caption{Interpolating only $3$ frames with a large deformation.}
\label{fig:interface} \end{figure}

\paragraph{Using fewer frames for large deformation.} To demonstrate our robustness, we interpolate the result of \cref{fig:raw_scans} Bottom using just $3$ frames (\cref{fig:interface}), maintaining robustness. To compare with source–target shape deformation, due to unavailable code for 4Deform \cite{4Deform} and issues with training scripts and pretrained models in Implicit-Surf-Deformation (ISD) \cite{sang2025implicit}, we adopt Neural Implicit Surface Evolution (NISE) \cite{novello2023neural}, which is a common SOTA baseline under the same experimental settings (by implicitly fitting two shapes and building pairwise correspondences). We also compare to reconstructing the three frames by Poisson reconstruction which is sensitive to noise and outliers, and generating correspondence via Unsupervised Learning of Robust Spectral Shape Matching \cite{cao2023unsupervised}. The result is visibly poor, not meeting the prerequisites for methods \cite{eisenberger2019divergence,cao2024spectral}.

\begin{figure}[!htb]
\centering
\includegraphics[width=1.\linewidth]{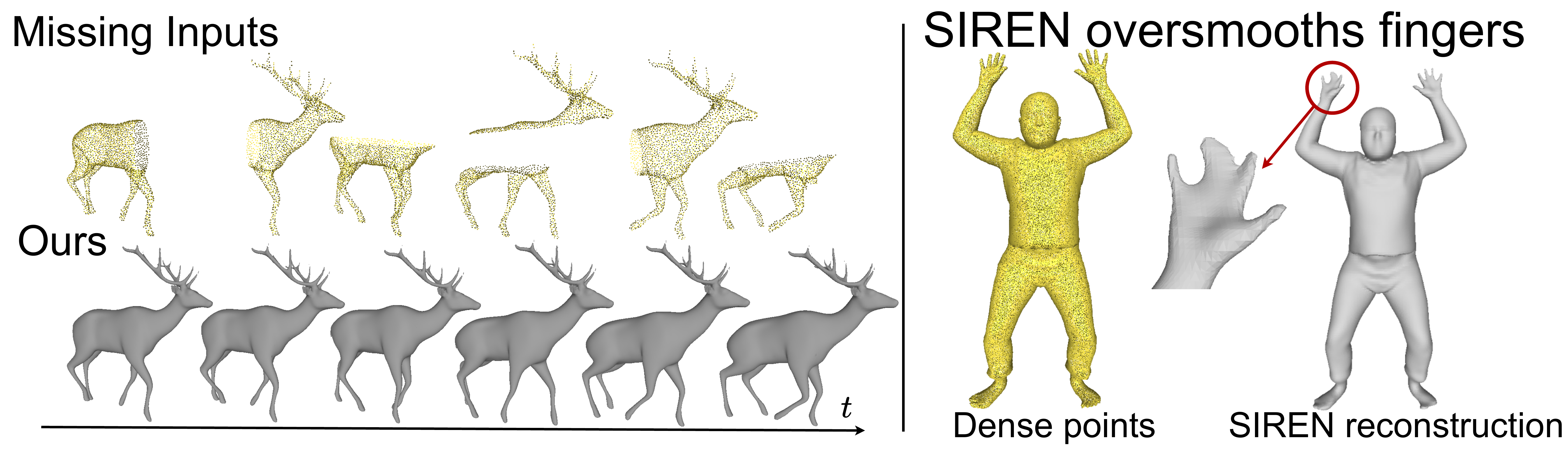}
\caption{Left: Ours restores regions; Right: SIREN oversmooths.}
\label{fig:two}
\end{figure}

\paragraph{Missing Regions.} We reran (\cref{fig:two}) the experiment from \cref{fig:miss},  using point clouds with significantly more missing regions ratios (ratios 0.35--0.63, mean 0.47). Even under these extreme conditions, our method completes well and remains robust.

\paragraph{Loss of Details.} SIREN-based implicit functions tend to oversmooth fine structures like fingers. We show this in an ``upper-bound'' static reconstruction with the same number of samples as in \emph{the entire sequence} (\cref{fig:regular_sparse} and \cref{fig:two} Right). Mitigating this oversmoothing is beyond our scope as a valuable direction for future work.
\clearpage

\end{document}